\documentclass[letterpaper, 10 pt, conference]{ieeeconf}  %

\usepackage[utf8]{inputenc} %
\usepackage[T1]{fontenc}    %
\usepackage{url}            %
\usepackage{booktabs}       %
\usepackage{amsfonts}       %
\usepackage{nicefrac}       %
\usepackage{microtype}      %
\usepackage{xcolor}         %
\usepackage{pifont}
\usepackage{tabularx}
\usepackage{array}
\let\labelindent\relax
\usepackage{enumitem}
\usepackage{graphicx}
\usepackage{float}
\usepackage{wrapfig}
\usepackage{caption, subcaption, overpic, textpos}
\usepackage{makecell}
\usepackage{comment}
\usepackage{amsmath}
\usepackage{amssymb}
\usepackage{multicol,multirow}
\usepackage{xspace}
\usepackage{lipsum}

\makeatletter
\DeclareRobustCommand\onedot{\futurelet\@let@token\@onedot}
\def\@onedot{\ifx\@let@token.\else.\null\fi\xspace}

\def\eg{\emph{e.g}\onedot}

\def\ie{\emph{i.e}\onedot}

\makeatother

\usepackage[svgnames]{xcolor}
\usepackage[table]{xcolor}

\definecolor{green}{RGB}{0,150,10}
\definecolor{blue}{RGB}{0,148,181}
\definecolor{orange}{RGB}{194,153,107}
\definecolor{darkblue}{RGB}{0,50,195}
\definecolor{myblue}{rgb}{0.21,0.49,0.74} 

\renewcommand{\methodname}{FreeTacMan\xspace}   

\usepackage[table]{xcolor}  
\definecolor{tablehighlight}{HTML}{FFFFE0}   
\definecolor{lightgraycolor}{HTML}{E5E5E5}  

\usepackage{ifpdf}  
\ifpdf

  \newcommand{\lightgray}{\cellcolor{lightgraycolor}}
\else

  \newcommand{\lightgray}{}
\fi
\usepackage{makecell} 
\usepackage{graphicx}
\usepackage{multirow}
\usepackage{xspace}
\usepackage{url}
\usepackage[breaklinks,colorlinks,linkcolor=red,urlcolor=magenta,citecolor=cvprblue]{hyperref}
\usepackage{makecell} 
\usepackage{mathtools} 
\usepackage{capt-of}
\usepackage{stfloats}  
\usepackage{cuted}     

\usepackage[capitalise]{cleveref}
\crefname{section}{Sec.}{Secs.}
\Crefname{section}{Section}{Sections}
\Crefname{figure}{Figure}{Figures}
\crefname{figure}{Fig.}{Figs.}
\Crefname{table}{Table}{Tables}
\crefname{table}{Tab.}{Tabs.}

\IEEEoverridecommandlockouts                              %

\title{\LARGE \bf
\methodname: Robot-free Visuo-Tactile \\
Data Collection System for Contact-rich Manipulation
}

\author{
    \textbf{Longyan Wu} $^{1,4*}$ \quad
    \textbf{Checheng Yu} $^{2*}$ \quad
    \textbf{Jieji Ren} $^{3*}$ \quad
    \textbf{Li Chen} $^{2}$ \quad
    \textbf{Yufei Jiang} $^{5}$ \\
    \textbf{Ran Huang} $^{4}$ \quad 
    \textbf{Guoying Gu} $^{3}$ \quad
    \textbf{Hongyang Li} $^{2,1}$ \\
    $^{1}$ Shanghai Innovation Institute \quad
    $^{2}$ The University of Hong Kong \\ 
    $^{3}$ Shanghai Jiao Tong University \quad
    $^{4}$ Fudan University \quad
    $^{5}$ Shanghai University \\
    \url{https://opendrivelab.com/FreeTacMan}
}

\begin{document}
\maketitle
\thispagestyle{empty}
\pagestyle{empty}

\ifdefined\HCode
  \begin{figure*}[!ht]
  \centering
  \includegraphics[width=\textwidth]{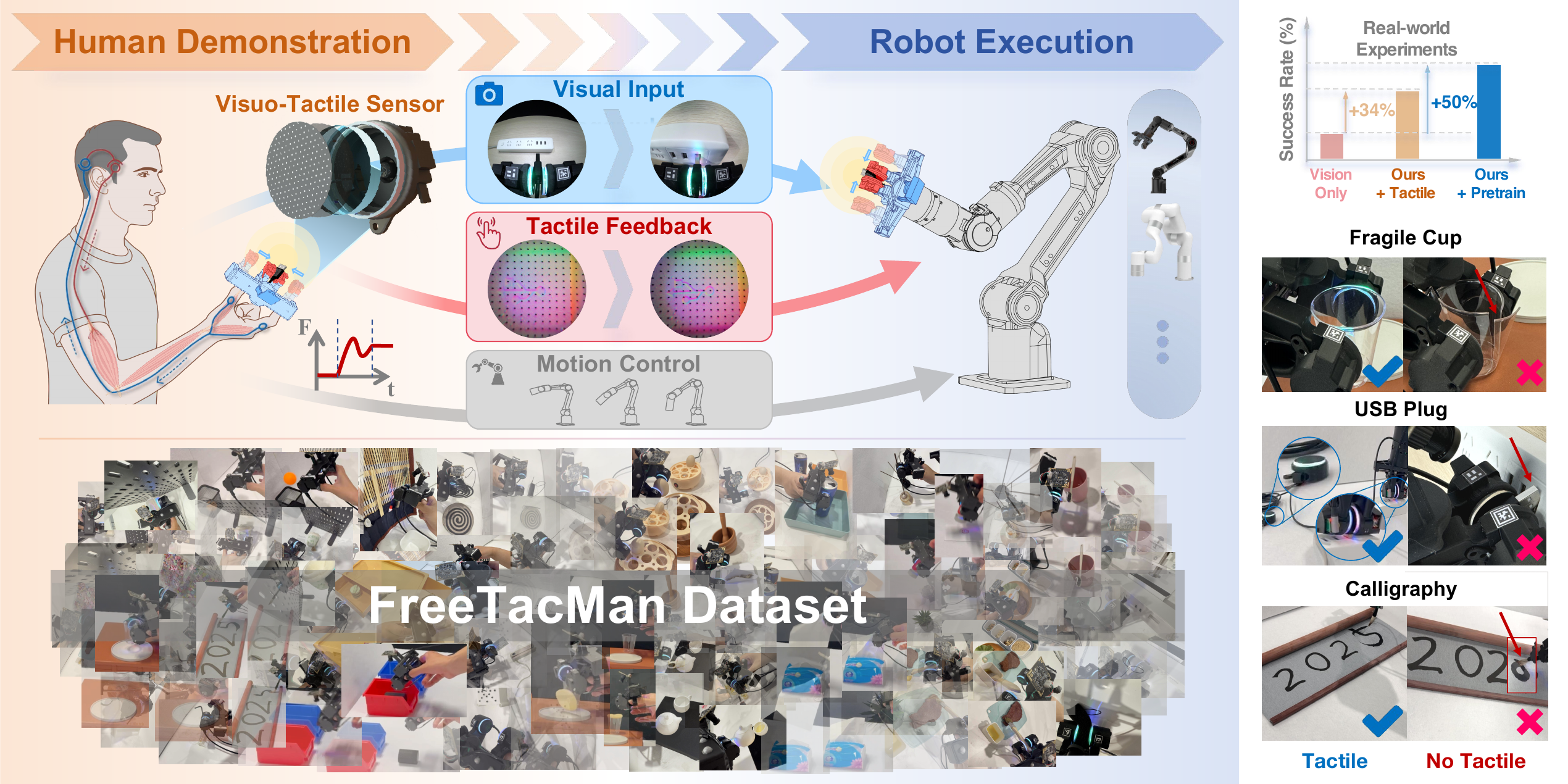}%
  \caption{%
      \textbf{Overview of \methodname}. \methodname is a robot-free, human-centric visuo-tactile data collection system that enables the efficient transfer of human visual, tactile, and motor skills to robots. It facilitates the collection of large-scale, contact-rich manipulation datasets.
      Project page: \url{https://opendrivelab.com/FreeTacMan}.%
  }
  \label{fig:teaser}
  \end{figure*}
\else
  \begin{strip}
  \vspace{-5em}
  \centering
  \begin{minipage}{\textwidth}
    \centering
    \includegraphics[width=\textwidth]{imgs/FreeTacMan_teaser2.pdf}%
    \captionof{figure}{%
      \textbf{Overview of \methodname}. \methodname is a robot-free, human-centric visuo-tactile data collection system that enables the efficient transfer of human visual, tactile, and motor skills to robots. It facilitates the collection of large-scale, contact-rich manipulation datasets.
      Project page: \url{https://opendrivelab.com/FreeTacMan}.
    }\label{fig:teaser}
  \end{minipage}%
  \end{strip}
\fi

\begin{abstract}

    Enabling robots with contact-rich manipulation remains a pivotal challenge in robot learning, which is substantially hindered by the data collection gap, including its inefficiency and limited sensor setup. While prior work has explored handheld paradigms, their rod-based mechanical structures remain rigid and unintuitive, providing limited tactile feedback and posing challenges for operators. Motivated by the dexterity and force feedback of human motion, we propose \methodname, a human-centric and robot-free data collection system for accurate and efficient robot manipulation. Concretely, we design a wearable gripper with visuo-tactile sensors for data collection, which can be worn by human fingers for intuitive control. A high-precision optical tracking system is introduced to capture end-effector poses while synchronizing visual and tactile feedback simultaneously. We leverage \methodname to collect a large-scale multimodal dataset, comprising over 3000k paired visuo–tactile images with end-effector poses, 10k demonstration trajectories across 50 diverse contact-rich manipulation tasks. 
    \methodname achieves multiple improvements in data collection performance over prior works and enables effective policy learning from self-collected datasets.
    By open-sourcing the hardware and the dataset, we aim to facilitate reproducibility and support research in visuo-tactile manipulation.

\end{abstract}

\section{INTRODUCTION}

Humans inherently rely on the integration of vision and touch to perform contact-rich manipulation tasks. While vision provides comprehensive object recognition and pose estimation capabilities, tactile feedback conveys critical information about local contacts that cannot be obtained visually, such as surface texture~\cite{connor1992neural}, in-hand pose \cite{xu2023visual}, material compliance~\cite{tiest2009cues}, and force distribution~\cite{ma2019dense}. 
For example, when handling a fragile or deformable object, vision guides initial motion planning and object localization, while tactile information enables modulation of grip force and adjustment of object orientation to prevent damage. 
The vision-based imitation learning has shown strong potential in robot manipulation tasks~\cite{yang2026riseselfimprovingrobotpolicy,shi2026egohumanoidunlockinginthewildlocomanipulation,shi2025diversity,jiang2025wholebodyvla}, benefiting from the growing large-scale demonstration datasets~\cite{padalkar2023open, bu2025agibot}. 
In the tactile domain, visuo-tactile sensors offer high-resolution, sensitive, and easily integrable multimodal signals, making them particularly well-suited for existing policy learning pipelines.
However, the lack of large-scale, high-quality tactile datasets and compatible sensing hardware has prevented similar advances. 
To accelerate research in this domain, two prerequisites would be addressed: 
1) a visuo-tactile data collection system that provides real-time tactile feedback and enables rapid redeployment, and 2) a large-scale, high-precision visuo-tactile dataset covering diverse contact-rich manipulation tasks~\cite{chen2025intelligent}.

Early efforts concentrated on collecting data using sensors mounted on robots~\cite{lin2025learning, huang20243d},
which limit the flexibility for scaling visuo-tactile data. 
Systems that rely on motion-capture with AR/VR visualization~\cite{xue2025reactive} or primary-replica teleoperation rigs~\cite{fu2024mobile} capture accurate camera views and trajectories.
They conversely offer no direct, real-time tactile signals, and impose a fixed robot setup with complex calibration or high latency. 
Handheld data collection paradigms~\cite{chi2024universal, zhaxizhuoma2025fastumi} release the human operators from robot embodiment. However, these methods typically rely on SLAM and IMU fusion for localization, which introduces significant positioning errors that are particularly detrimental to tactile perception. Moreover, they transmit tactile cues through long mechanical linkages, hindering direct tactile feedback and precise measurement of instantaneous tactile signals.

The inefficiency and lack of real-time tactile feedback in current data collection setups compromise data quality, limit scalability, and pose a risk of sensor damage during operation. Consequently, existing visuo-tactile datasets~\cite{dou2024tactile,cheng2025touch100k} are largely confined to specific perception tasks or lack the diversity and precision required for generalizable visuo-tactile policy learning across diverse manipulation scenarios.

In this work, we introduce \textbf{\methodname}, 
a robot-free and human-centric visuo-tactile data collection system to acquire manipulation data accurately and efficiently.
With a modular sensor and in-situ\footnote{``In-situ'' indicates that~\methodname preserves natural fingertip–environment interaction. It emphasizes maintaining the original grasp posture while capturing tactile feedback. This is akin to the concept in materials science~\cite{tao2011situ} and biology~\cite{zheng2024high}, where systems are studied in their native conditions—without altering their environment or state—to gain a deeper understanding of their behavior under actual working conditions.} 
hardware, it ensures precise, real-time tactile feedback via two mechanisms - a gripper-finger interface coupled to human fingertips, and a linear transmission mechanism enabling precise gripper control and unattenuated tactile feedback.
This foundation enables a comprehensive visuo-tactile manipulation dataset with high precision, diverse task scenarios, and rich contact dynamics, recorded with synchronized high-resolution visual, visuo-tactile, and pose data. This extensive collection provides a robust foundation for training and evaluating generalizable visuo-tactile policies.
To validate the effectiveness of the
data collected by \methodname, we train and deploy imitation learning policies that leverage a temporal-aware tactile pretraining strategy on 
challenging manipulation tasks. 
In summary, our main \textbf{contributions} are: 

\textbf{(i)}
An in-situ, robot-free, real-time tactile data-collection \textit{system} that leverages a handheld gripper with modular visuo-tactile sensors to excel at 
diverse contact-rich tasks efficiently.

\textbf{(ii)} 
A large-scale, high-precision (sub-millimeter) visuo-tactile manipulation \textit{dataset} with over 3000k visuo-tactile image pairs, more than 10k trajectories across 50 tasks.

\textbf{(iii)}
Experimental validation shows that imitation policies trained with our visuo-tactile data achieve an average 50\% higher success rate than vision-only approaches across
challenging 
contact-rich manipulation tasks.

\begin{table}[t]
  \centering
  \caption{\textbf{Comparison with existing data collection systems.} 
    Our in-situ design delivers direct and high-precision tactile feedback to operators. To evaluate tactile feedback fidelity of handheld methods quantitatively, we measure the number of mechanical transmission ``link'' inside the handheld gripper from the human hand to the grasped object.
  }
  \label{tab:comparison}
  \vspace{3pt}
  \scalebox{0.9}{
  \begin{tabular}{llcc}
    \toprule
    \textbf{Category}  & \textbf{Method}      & \textbf{Control Method}     & \textbf{Tactile Feedback} \\
    \midrule
    \multirow{4}{*}{\makecell[l]{Teleop:\\VR/AR}}
      & ARCap \cite{chen2024arcap}           & VR Controller         & –                     \\
      & DexCap \cite{wang2024dexcap}         & Hand Mocap            & –                     \\
      & TactAR \cite{xue2025reactive}        & VR Controller         & Visual                \\
      & Bunny-VisionPro \cite{bunny-visionpro} & Hand Retargeting & Vibration            \\
    \midrule
    \multirow{3}{*}{\makecell[l]{Teleop:\\Primary–\\Replica}}
      & ALOHA \cite{fu2024mobile}            & Puppet Arm            & –                     \\
      & GELLO \cite{wu2024gello}             & Puppet Arm            & –                     \\
      & Bi-ACT \cite{buamanee2024bi}         & Puppet Arm            & Force                 \\
    \midrule
    \multirow{2}{*}{\makecell[l]{Handheld}}        
      & UMI~\cite{chi2024universal}, FastUMI~\cite{zhaxizhuoma2025fastumi} & \multirow{2}{*}{\makecell{Trigger}} & \multirow{2}{*}{\makecell{Contact (4 links)}} \\
      & ViTaMIn~\cite{liu2025vitamin} &  & \\
    \midrule
    \ifpdf
   
      \rowcolor{yellow!20}
      \textbf{In-situ} & \textbf{\methodname (Ours)} & Fingertips & Touch (1 link) \\
    \else

      \textbf{In-situ} & \textbf{\methodname (Ours)} & Fingertips & Touch (1 link) \\
    \fi
    \bottomrule
  \end{tabular}
  }
\vspace{-10pt}
\end{table}

\section{Related Work}
\subsection{Dataset Collection System for Robot Learning.}

Recent robot learning has been greatly advanced through imitating expert demonstrations~\cite{pan2025agility}. Popular approaches~\cite{zhao2023act,chi2023diffusion} leverage large-scale visuomotor datasets (pairs of RGBD sensors and actions) to train end-to-end policies that generalize across objects and scenes. 
To gather data suitable for imitation learning, prior works have relied on different interfaces to teleoperate real robots, including motion capture with AR or VR visualization~\cite{chen2024arcap, bu2025agibot, xue2025reactive}, 3D space mouse~\cite{chi2023diffusion}, primary-replica system~\cite{fu2024mobile, wu2024gello, buamanee2024bi}, and wearable devices~\cite{brygo2014humanoid}. However, these setups are either expensive, operationally complex, or suffer from limited precision.
Handheld data collection paradigms~\cite{chi2024universal} offer greater flexibility for various robot embodiments, but their trigger-based grippers and multi-link transmission designs introduce backlash between links, which blurs tactile cues and leads to inaccurate relay of gripper contact to the operator.
In contrast, \methodname implements an in-situ data collection system that enables operators to feel gripper contact directly in real time. The comparison of control methods and tactile feedback between ours and prior work is depicted in~\Cref{tab:comparison}.

\subsection{Tactile dataset for contact-rich Manipulation.}

Most existing visuo-tactile datasets focus on perception tasks such as 6DoF pose tracking~\cite{Huang_2025}, cross-modal generation~\cite{dou2024tactile}, representation learning~\cite{cheng2025touch100k}, 
garment-feature spatially-aligned perception~\cite{DBLP:conf/rss/KerrHWHICG23}, and tactile-semantic description~\cite{fu2024touchvisionlanguagedataset}, with limited coverage of the full manipulation process.
The ObjectFolder benchmark~\cite{gao2023objectfolder} includes 4 visuo-tactile manipulation tasks, exhibiting limited scale and diversity.
Moreover, most tactile manipulation datasets are collected with tactile array sensors based on piezoresistive or Hall-effect principles~\cite{huang20243d,10710144, liu2025vtdexmanip, zhu2025touchwildlearningfinegrained}. Although policies trained on these datasets show promising results, the intrinsic limitations of these sensors, such as low spatial resolution, crosstalk, complex fabrication, and environment interference, can constrain policy performance and reduce deployment efficiency in real-world settings.
Taken together, these limitations expose an urgent demand for a visuo-tactile dataset that offers rich, high-resolution, and scalable data spanning the entire manipulation process.

\section{Method}
\label{sec:method}

\subsection{Hardware Design}
\label{sec:method-hardware}

   \begin{figure*}[t]
      \centering
      \includegraphics[width=\linewidth]{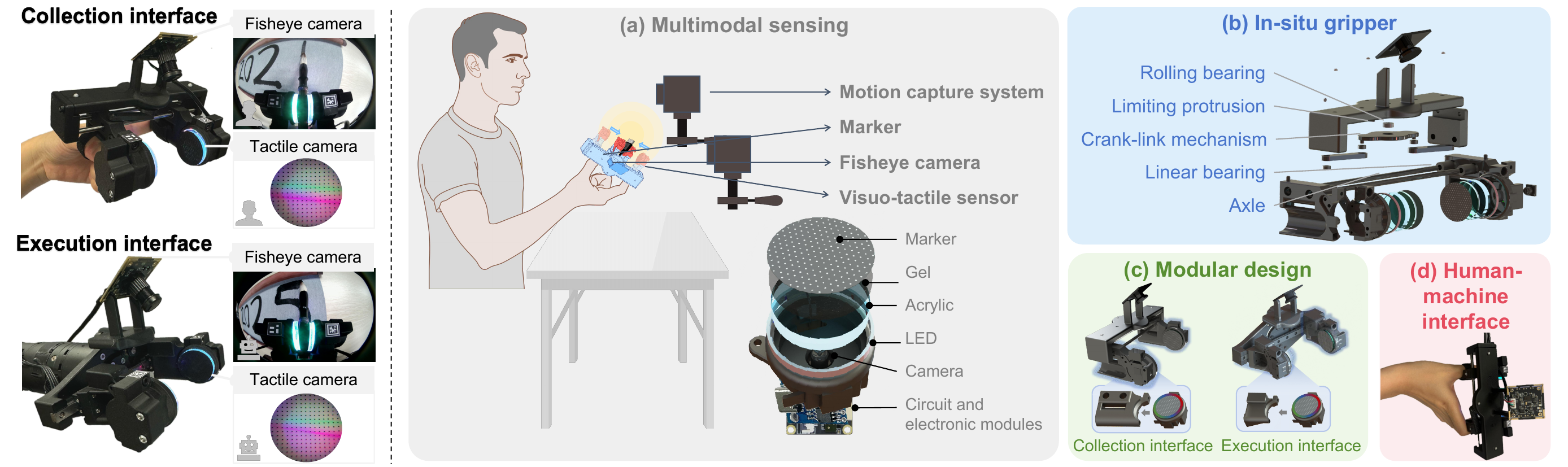}
      \caption{\textbf{Hardware system}. \textit{Left}: The in-situ gripper in the collection and execution interface respectively, with identical visual and tactile observations. \textit{Right}: (a) Composition of the sensor. (b) Exploded view of 
    \methodname.
    (c) The modular design allows for an agile switch
    between the collection and execution interface. (d) Human-machine interface design.}
      \label{fig:hardware}
      \vspace{-10pt}
   \end{figure*}
\textbf{Design criteria.}
To enable efficient collection of high-fidelity tactile data, we define the following criteria.
\textbf{(a)} Multimodal data acquisition: 
The system must provide competitive visuo-tactile sensing with exceptional consistency, coupled with high-precision pose tracking.
\textbf{(b)} Efficiency: 
The system should minimize the tactile transmission path from human fingers to the grasped object for real-time and precise tactile feedback, while ensuring stable and fingertip-level control. 
\textbf{(c)} Scalability: A modular design architecture should be adopted to ensure compatibility across robot embodiments.
\textbf{(d)} Usability: The system should accommodate a wide range of fingertip sizes, providing ergonomic comfort during operation. 

\textbf{Multimodal sensing.}
As shown in~\cref{fig:hardware}(a), we employ visuo-tactile sensors and a wrist-mounted camera to capture tactile and visual information. 
End-effector poses are tracked using a high-precision motion capture system, achieving sub-millimeter accuracy—an essential factor for contact-rich tasks, where even minute pose errors can result in disproportionately large deviations in tactile feedback. 

\textbf{In-situ gripper.}
\methodname achieves hinge-free operation through visuo-tactile sensors mounted on the operator's fingertip, where the sensor layer forms the interface between skin and manipulated objects, eliminating intermediate linkages to provide zero mechanical attenuation and natural proprioception. To ensure that the unattenuated tactile feedback directly translates to operator control precision, as illustrated in~\cref{fig:hardware}(b), the system incorporates a linear transmission mechanism with chrome-plated steel shafts and linear bearings, constraining movement to highly accurate linear trajectories (axial deviation $\geq$ 0.02 mm). Additionally, an inverted crank-slider mechanism converts finger-driven motion into synchronized linear output, while dual parallel shafts and rolling bearings in linking joints minimize friction and lateral torque, achieving over 90\% transmission efficiency.

\textbf{Modular architecture.}
\methodname system is built as three plug-and-play modules, each optimized for rapid setup and cross-embodiment compatibility:
a sensor perception module for tactile data collection, 
a universal gripper interface (\cref{fig:hardware}(c))
for robot compatibility, and 
a camera mounting scaffold to ensure aligned visual feedback from wrist camera.
The sensor is based on the McTac design~\cite{DBLP:conf/icira/RenZG23} and features enhanced modularity and consistency. 
The improved universal mechanical interface allows the same sensor unit to be used interchangeably on a data collection setup or a robotic arm. Automated fabrication ensures consistent quality.
Customized interfaces are provided for different robot arms, including a 6-DOF Piper arm for low-load tasks and a 7-DOF Franka arm for high-precision, heavy-load applications. 
For 3D models of end-effectors and their integration on different robot arms (\textit{e.g.}, Piper, Franka), please visit \href{https://opendrivelab.com/FreeTacMan}{our demo page}. 

\textbf{Human-machine interface design.}
To achieve rapid adaptability, we incorporate hook-and-loop straps for fingertip fastening, as illustrated in~\cref{fig:hardware}(d).
These straps fit hand sizes ranging from the 5\% to 95\% percentile of adults and support repeated use~\cite{zoeller2020systematic}, balancing operational efficiency. 
Furthermore, the system features a lightweight ($157.5,g$) and compact ($145 \times 85 \times 106,\text{mm}^3$) design that ensures comfort.

\subsection{Human-to-Robot Data Transfer }
\label{sec:method-data}
To facilitate human-to-robot skill transfer, a high-precision NOKOV Motion Capture System is used for 6D pose tracking of the interface at 240 Hz, achieving sub-millimeter accuracy. 
Five retro-reflective markers are mounted on the interfaces, with three positioned on the top plate to measure pose and two on the grippers to capture relative displacement.
The coordinates of markers are first transformed from the world coordinate frame to the robot base coordinate frame. A local coordinate frame, aligned with the robot URDF, is established at the gripper’s Tool Center Point (TCP) using three top-plate markers, allowing derivation of end-effector pose while ensuring consistency with the robot kinematic model.
We downsample the tracking data to synchronize with RGB images.
To this end, each frame contains: the wrist camera RGB image, two visuo-tactile images, the end-effector pose of the in-situ gripper in the world coordinate frame, and gripper width. Each trajectory consists of a sequence of such embodiment-agnostic data synchronized at 30 Hz.

Unlike prior works ~\cite{chi2024universal, zhaxizhuoma2025fastumi} 
relying on SLAM
and IMU fusion to estimate
end-effector poses, the motion capture system avoids IMU drift and tracking errors. Given the URDF of a target embodiment, we could utilize IKPY~\cite{Manceron_IKPy} as an inverse kinematic solver to map poses of the in-situ gripper to joint positions 
directly as the action representation.

\subsection{Tactile Pretraining and Policy Learning}
\label{sec:method-policy}
A two-stage approach is employed to learn visuo-tactile manipulation policies: 1) Tactile representation learning, 2) visuo-tactile policy learning.

   \begin{figure*}[t]
      \centering
      \includegraphics[width=\linewidth]{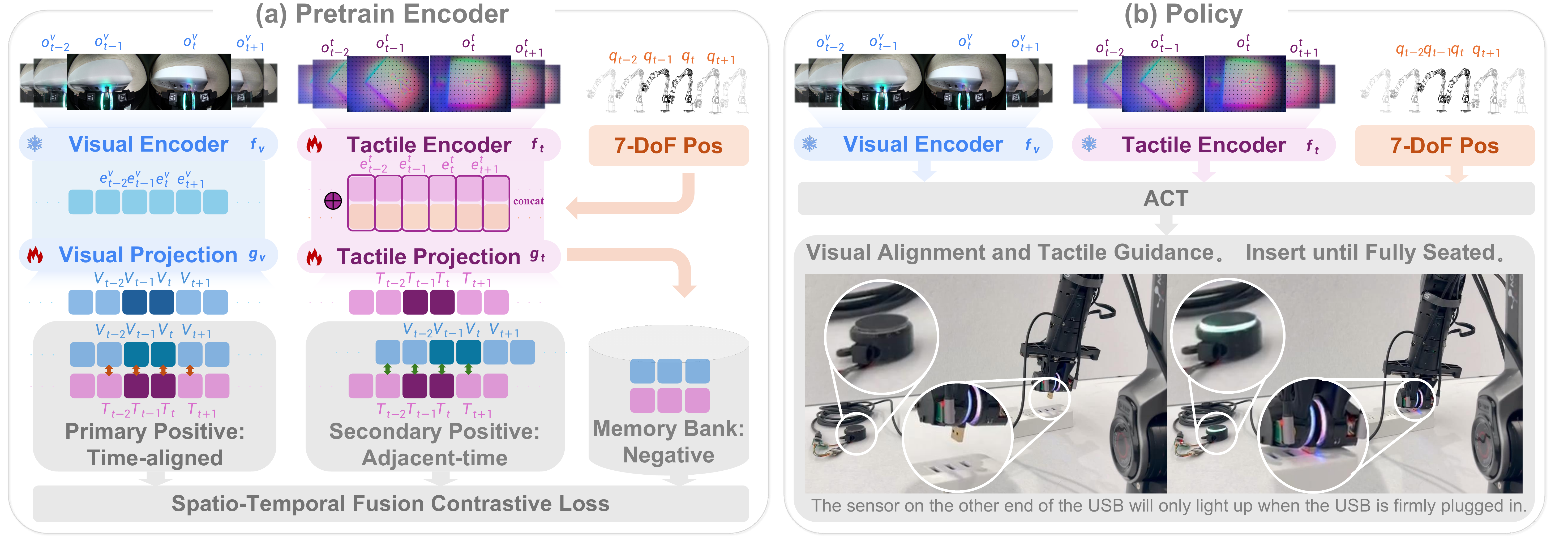}
      \caption{\textbf{Tactile pretraining and policy learning pipeline.} (a) A tactile encoder is pretrained using the self-collected dataset. (b) The pretrained tactile encoder is integrated into an ACT-based policy for downstream tasks such as USB insertion.}
      \label{fig:policy}
      \vspace{-10pt}
   \end{figure*}

\textbf{Tactile representation learning.}
Our wearable system yields high-precision, contact-rich visuo-tactile trajectories, enabling a high-fidelity dataset for representation learning. Although tactile outputs resemble 2D images, applying a vision encoder pretrained on RGB data often produces suboptimal features due to the domain gap in appearance and semantics~\cite{george2024visuo, liu2025vitamin}.
We adopt a CLIP-style~\cite{radford2021learning} contrastive pretraining procedure to bridge the domain gap.

As illustrated in~\cref{fig:policy}(a), both the visual encoder $f_v$ and tactile encoder $f_t$ share a ResNet backbone initialized from the same checkpoint. $f_v$ remains frozen during pretraining, while $f_t$ is finetuned.
Each encoder is followed by a projection head: $g_v$
for vision, and $g_t$ for tactile, where $g_t$ first concatenates the tactile features with the normalized 7-DOF joint position vector $\mathbf{q}_i$ to inject robot joint state as global context.
At each timestep $i$, we compute the normalized embeddings 
$\mathbf{v}_i$
and $\mathbf{t}_i$.
We designate \(\mathbf{v}_i\) as the \emph{primary positive} for $\mathbf{t}_i$ to align tactile features with the current visual features.
However, relying solely on time-aligned contrastive loss neglects temporal dynamics, leading to embeddings that vary abruptly and fail to capture evolving contact patterns. 
To enforce temporal awareness, a \emph{secondary positive} \(\mathbf{v}_{i+1}\) drawn from the next timestep is introduced.
All other entries from negatives in a fixed-size memory bank $\mathcal{{M}}$. We train $f_t$, $g_v$ and $g_t$ by minimizing the following contrastive loss:

\begin{equation}
L = -\frac{1}{B}\sum_{i=1}^{B}
\log
\frac{
    e^{\mathbf{v}_i^\top \mathbf{t}_i / \tau}
  + e^{\mathbf{v}_{i+1}^\top \mathbf{t}_i / \tau}
}{
    e^{\mathbf{v}_i^\top \mathbf{t}_i / \tau}
  + e^{\mathbf{v}_{i+1}^\top \mathbf{t}_i / \tau}
  + \sum_{j \in \mathcal{N}_i} e^{\mathbf{v}_j^\top \mathbf{t}_i / \tau}
},
\end{equation}
where $B$ indicates batch size, $\tau$ is a learned temperature parameter and \(\mathcal{N}_i\) indexes the negatives. 
\textbf{Visuo-tactile action chunking transformer.}
We employ the pretrained tactile encoder to extract tactile representations. As shown in~\cref{fig:policy}(b), vision and tactile embeddings are then concatenated and input into the action chunking transformer (ACT)~\cite{zhao2023act}, which is trained to predict joint positions. 

\section{Experiments}
\label{sec:exp}

We design experiments to answer three key questions: 

\textbf{Q1}. Can demonstrations be collected efficiently and accurately using \methodname compared to previous setups?

\textbf{Q2}. Is the in-situ tactile information in \methodname dataset effective for contact-rich tasks policy learning?

\textbf{Q3}. How does the tactile encoder pretrained on self-collected visuo-tactile data improve policy learning? 

\subsection{Dataset}
\label{sec:method-dataset}
   \begin{figure*}[t]
      \centering
      \includegraphics[width=\textwidth]{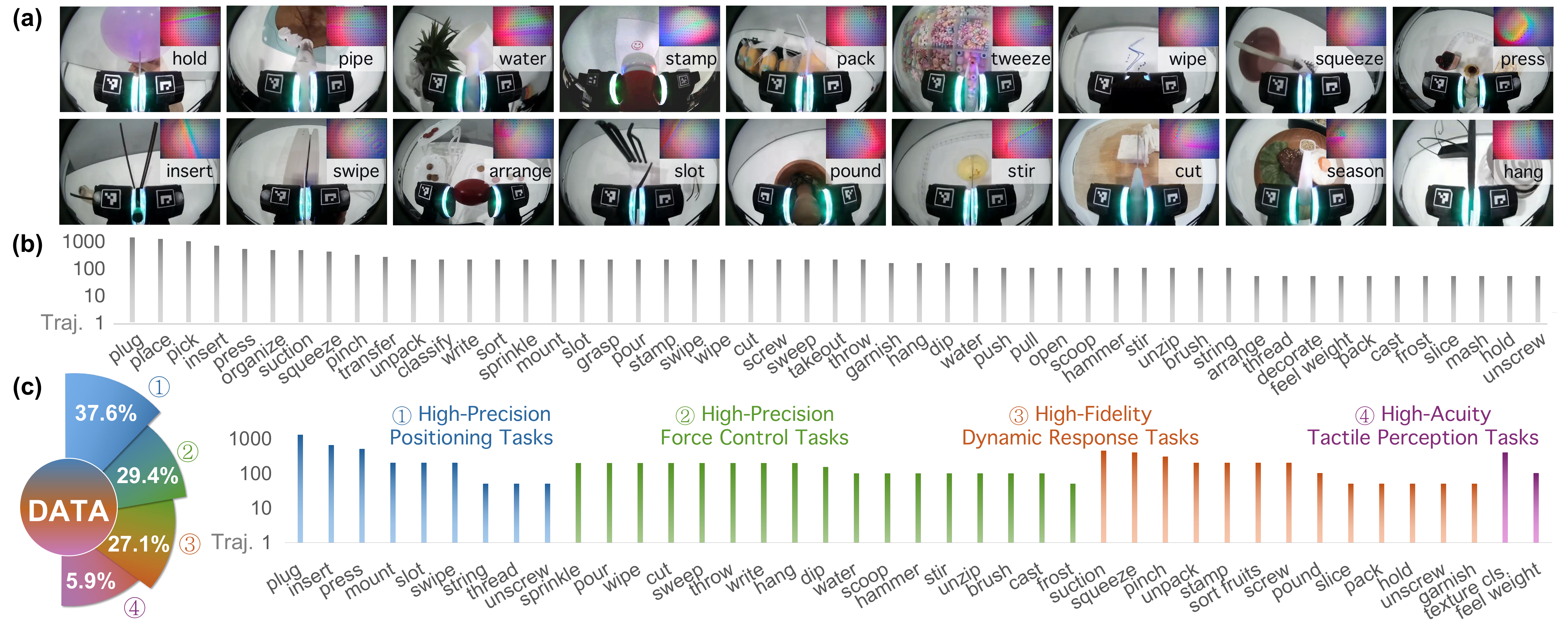}
      \caption{\textbf{The \methodname dataset.} 
      (a) Representative examples illustrating the diversity in both task complexity and tactile context. (b) The dataset covers 50 tasks and features a large-scale collection of data, including more than 10k trajectories and over 3000k visuo-tactile pairs. (c) The dataset enables diverse fundamental tactile capabilities.
      }
      \label{fig:datasetall}
      \vspace{-10pt}
   \end{figure*}

Enabled by the efficient, precise, and faithful tactile data collection system, we curate a diverse dataset spanning vision, touch, and proprioception modalities, as illustrated in \cref{fig:datasetall}. 
The dataset spans 50 tasks, comprising more than 10k trajectories and over 3000k visuo-tactile image pairs. Collected with high-consistent and high-resolution visuo-tactile sensors, it enables large-scale tactile pretraining and multimodal policy learning. Furthermore, the dataset covers diverse fundamental tactile capabilities, as shown in~\cref{fig:datasetall}(c).
To ensure dataset quality, every trajectory is replayed in our validation process, providing a robust filter that guarantees the reliability of the data for downstream robot learning. The dataset, along with its structure, data format, and license, is available at~\href{https://huggingface.co/datasets/OpenDriveLab/FreeTacMan}{our dataset page}.

\subsection{Experimental Setup}
\label{sec:exp-setup}

   \begin{figure}[t]
      \centering
      \includegraphics[width=\linewidth]{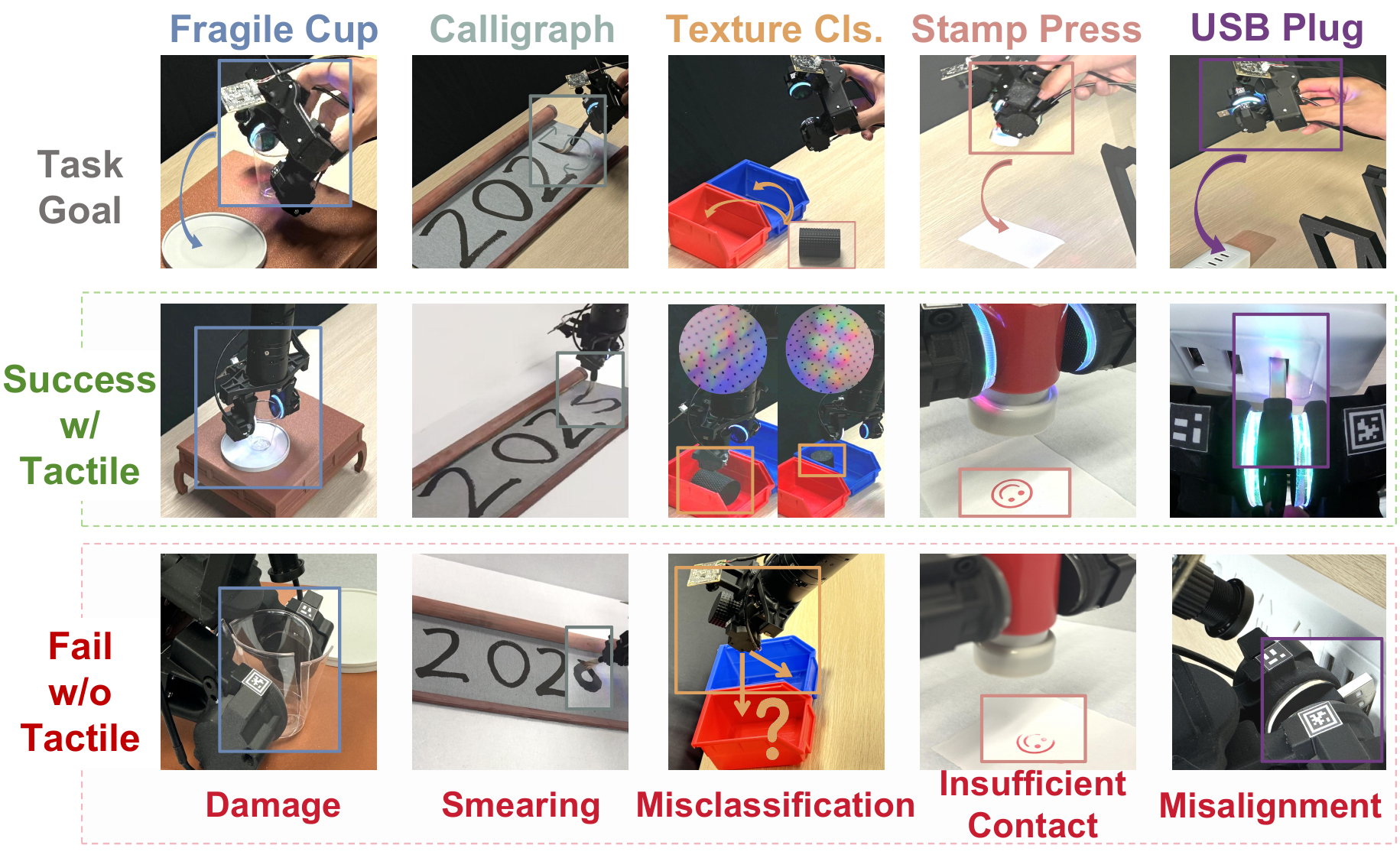}
      \caption{\textbf{Human demonstrations and policy rollouts.} The top row shows goal trajectories, the middle row demonstrates successful rollouts with tactile feedback, and the bottom row showcases typical failure modes without tactile input.}
      \label{fig:dataset}
      \vspace{-10pt}
   \end{figure}

To answer the questions above, we evaluate the effectiveness of \methodname system and the quality of the collected dataset through a diverse set of contact-rich manipulation tasks, shown in~\cref{fig:dataset}.
1) \textbf{Fragile cup}. The robot grasps a fragile cup without causing damage and places it stably on a tray. 2) \textbf{Calligraphy}. The robot traces the digit "5" with a calligraphy brush. 3) \textbf{Texture classification}. The robot grasps and identifies one of two cylindrical objects with distinct textures and sorts it into the correct bin. 4) \textbf{Stamp press}. The robot presses a stamp onto paper to produce a clear imprint. 5) \textbf{USB plug}. The robot needs to securely insert a pre-grasped USB plug into a socket (error $<1mm$).
These tasks collectively represent the four core tactile capabilities of our dataset: force control (fragile cup), hybrid force-position control (USB plug, stamp press), high-acuity tactile perception (texture classification), and dynamic response (calligraphy).

\subsection{User Study on Data Collection System}
\label{sec:exp-user}

   \begin{figure*}[t]
      \centering
      \includegraphics[width=\linewidth]{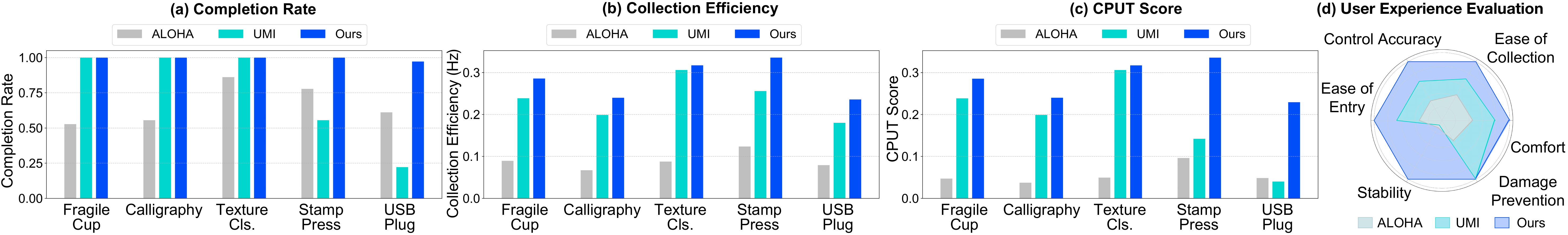}
      \caption{\textbf{
    User Study 
    on data collection.
    } (a-c) \methodname outperforms ALOHA~\cite{fu2024mobile} and UMI~\cite{chi2024universal}
    in terms of 
    completion rate, collection efficiency, and the 
    CPUT score per task.
    (d) \methodname excels at 
    user experience evaluation as well.}
      \label{fig:userstudy}
      \vspace{-5pt}
   \end{figure*}

   \begin{figure*}[t]
      \centering
      \includegraphics[width=\linewidth]{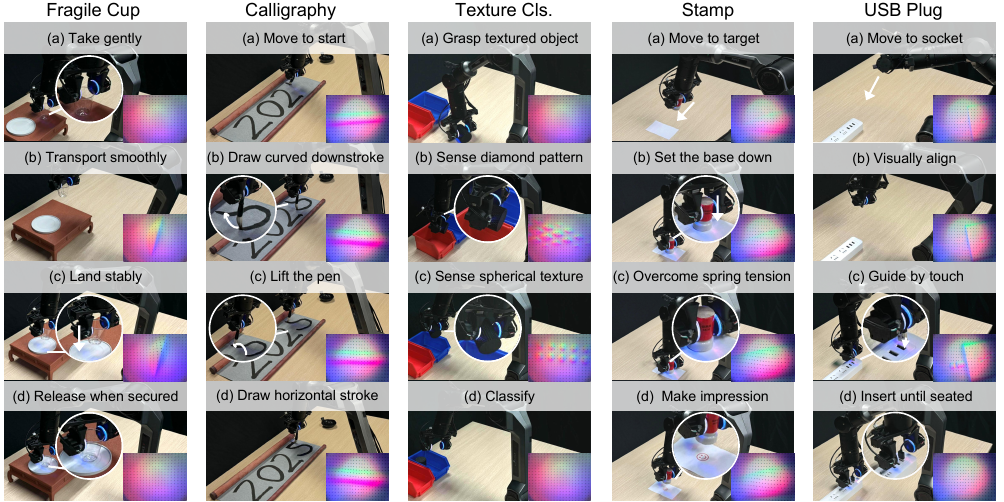}
      \caption{\textbf{Trajectory visualization.} We test \methodname on a variety of contact-rich tasks. Videos are available on  \href{https://opendrivelab.com/FreeTacMan}{our website}.}
      \label{fig:policyrollouts}
      \vspace{-10pt}
   \end{figure*}

\textbf{Procedure.}
We evaluate the usability of~\methodname through a user study including 12 volunteers, with varying experience in data collection. 
Besides~\methodname, users collect demonstrations using two typical methods: primary–replica-based teleoperation (\ie, ALOHA~\cite{fu2024mobile}) and handheld devices (\ie, UMI~\cite{chi2024universal}).
To minimize biases, 
no device-specific instructions are provided and each participant conducts three trials with each device for each task.
The participants are instructed to solve tasks as best they can while avoiding collisions and damage. If a task fails, they will continue from where it is interrupted, and the failure will be recorded.

\textbf{Metrics.}
We record the task success/failure mode, completion time, and any instances of slippage or damage. 
Three metrics are adopted to quantify the capability of a particular 
data collection approach,
namely $\texttt{completion\_rate}$ (fully completed tasks as a percentage of those initiated), collection $\texttt{efficiency}$ (the inverse of data collection time), and an overall score, Completion per Unit Time (\texttt{CPUT}), defined as $\texttt{completion\_rate} \times \texttt{efficiency}$. 
\textbf{Demonstrations could be collected efficiently and accurately using \methodname compared to previous setups (Q1).}
The results of user study are shown in~\cref{fig:userstudy}, where \methodname consistently yields the top completion rate and efficiency. CPUT score in~\cref{fig:userstudy}(c) highlights the overall advantage of \methodname, gaining 5.05$\times$ higher performance than teleoperation and 1.52$\times$ higher than UMI.

For simple tasks like texture classification that don’t require accurate force sensing or control, our system and UMI achieve comparable completion rates, while ALOHA performs slightly inferior. In completion time, we have a $2.65 \times$ advantage over ALOHA and a $1.22 \times$ advantage over UMI.
In more complex tasks, such as fragile cup manipulation, where force feedback is necessary to ensure successful grasping and prevent excessive force,
primary-replica teleoperation leads to damage-related failures. This suggests that the handheld method provides better force feedback for delicate objects.

For tasks requiring dynamic force control, such as USB plug and stamp press, ALOHA relies on brute force, barely completing them at the risk of causing damage, while UMI often fails due to the lack of slippage detection. In contrast, \methodname combines both precise force perception and control, resulting in superior accuracy and safety.
For high-precision trajectory control (\eg, calligraphy), ALOHA struggles significantly—users often need to manually assist the teaching arm with their other hand, resulting in the longest completion time. While UMI can complete the task, the trajectory smoothness remains inferior to that of \methodname.

\cref{fig:userstudy}(d) summarizes the user experience evaluation. Stability, comfort, ease of collection and entry are assessed via questionnaires and normalized. Stability and damage represent failures of object drops and damage. The results show that \methodname is the most user-friendly and reliable data collection system among the three approaches.

\begin{table}[t]
  \centering
  \caption{
    \textbf{Policy success rates (\%) across tasks.} Tactile input and pretraining significantly boost imitation learning.
  }
  \label{tab:ablation}
  \vspace{-2pt}
  \setlength{\tabcolsep}{3.5pt}  
  \scalebox{1}{
  \begin{tabular}{lcccccc}
    \toprule
    \textbf{Method} 
      & \makecell{Fragile\\Cup} & Calligraphy & \makecell{Texture\\Cls.} & \makecell{Stamp\\Press} & \makecell{USB\\Plug} & \lightgray\textbf{Avg.} \\
    \midrule
    \makecell{ACT~\cite{zhao2023act}\\(Vision-only)}      & 35 & 20 & 20 & 30 &  0 & \lightgray21 \\
    \makecell{Ours (+ Tactile \\w/o Pretrain)}          & 75 & 70 & 55 & 65 & 10 & \lightgray55 \\
    \makecell{Ours\\(+ Pretrain)}       & \textbf{80} & \textbf{90} & \textbf{85} & \textbf{80} & \textbf{20} & \lightgray\textbf{71} \\
    \bottomrule
  \end{tabular}
  \vspace{-18pt}
  }
\end{table}

\subsection{Validation on Imitation Learning}
\label{sec:exp-policy}

Each task in~\cref{fig:policyrollouts} is trained and evaluated over 20 trials.
\begin{itemize} [leftmargin=*,itemsep=0pt,topsep=0pt]
    \item \textbf{ACT}~\cite{zhao2023act} \textbf{(Vision-only)}: The original ACT uses RGB images from the wrist camera as input only. 
    \item \textbf{Ours (+ Tactile w/o Pretrain)}: An extended ACT model taking both visual and tactile observations, which are separately encoded by identical backbones without pretraining.
    \item \textbf{Ours (+ Pretrain)}: Our full model, where the tactile encoder is pretrained with a multi-positive contrastive objective incorporating both primary and secondary positives.
\end{itemize}

\textbf{Integration of tactile feedback results in significant improvements on policy performance (Q2).}
As illustrated in~\Cref{tab:ablation}, the vision-only baseline achieves low performance across all tasks, with an average success rate of 21\%. 
Without tactile feedback, the robot struggles in tasks requiring fine tactile perception, such as USB plug, calligraphy, and texture classification.
When tactile feedback is incorporated naively, \ie, without pretraining, performance improves substantially, with the success rate increasing to 55\%. Significant gains are observed in tasks such as fragile cup manipulation (75\%) and calligraphy (70\%).
This highlights the utility of tactile sensing in contact-rich tasks where visual cues alone are insufficient to distinguish subtle features. 
However, the performance in tasks like USB plug (10\%) and texture classification (55\%) still lags behind, indicating room for further refinement.

\textbf{Tactile inputs enable excellent performance even on unseen objects (Q2).} In the texture classification task, we select a texture object with an out-of-distribution color (red) for testing. As shown in \Cref{tab:unseen}, tactile input achieves 55\% accuracy on unseen objects, matching performance on training objects and significantly exceeding the 15\% vision-only baseline. Pretraining further increases accuracy to 70\%, demonstrating robust generalization.

\begin{table}[t]
\caption{\textbf{Comparison in unseen object.} Tactile input enables generalization to unseen objects.}
\label{tab:unseen}
\centering
\small 
\begin{tabular}{lccc}
\toprule
\textbf{} & \textbf{\makecell{Vision\\-only}} & \textbf{\makecell{Ours (+ Tactile \\w/o Pretrain)}} & \textbf{\makecell{Ours \\(+ Pretrain)}} \\
\midrule
training object & 20\% & 55\% & 85\% \\
\textbf{unseen object} & \textbf{15\%} & \textbf{55\%} & \textbf{70\%} \\
\bottomrule
\end{tabular}
\vspace{-10pt}
\end{table}

\textbf{Temporal-aware pretraining improves the performance by aligning visual and tactile embeddings (Q3).} 
Incorporating both time-aligned and time-adjacent pairs in CLIP pretraining boosts the average success rate to 71\%. 
As shown in the last line in~\Cref{tab:ablation}, the improvement comes from tasks that demand fine-grained control and the ability to track contact dynamics, such as calligraphy (90\%) and stamp (80\%). Meaningful gain is witnessed in USB plug (20\%). 
Nevertheless, this 20\% success rate highlights the challenge of high-precision tasks, where sub-millimeter deviations lead to failure. This is attributed to limited inverse kinematics accuracy and insufficient modeling of insertion dynamics. Future work will thus explore higher-precision arms and reinforcement learning to better learn contact-rich policies.
Overall, these results confirm that our approach aligns tactile and visual embeddings with spatial-temporal context.

\begin{figure}[t]
      \centering
      \includegraphics[width=\linewidth]{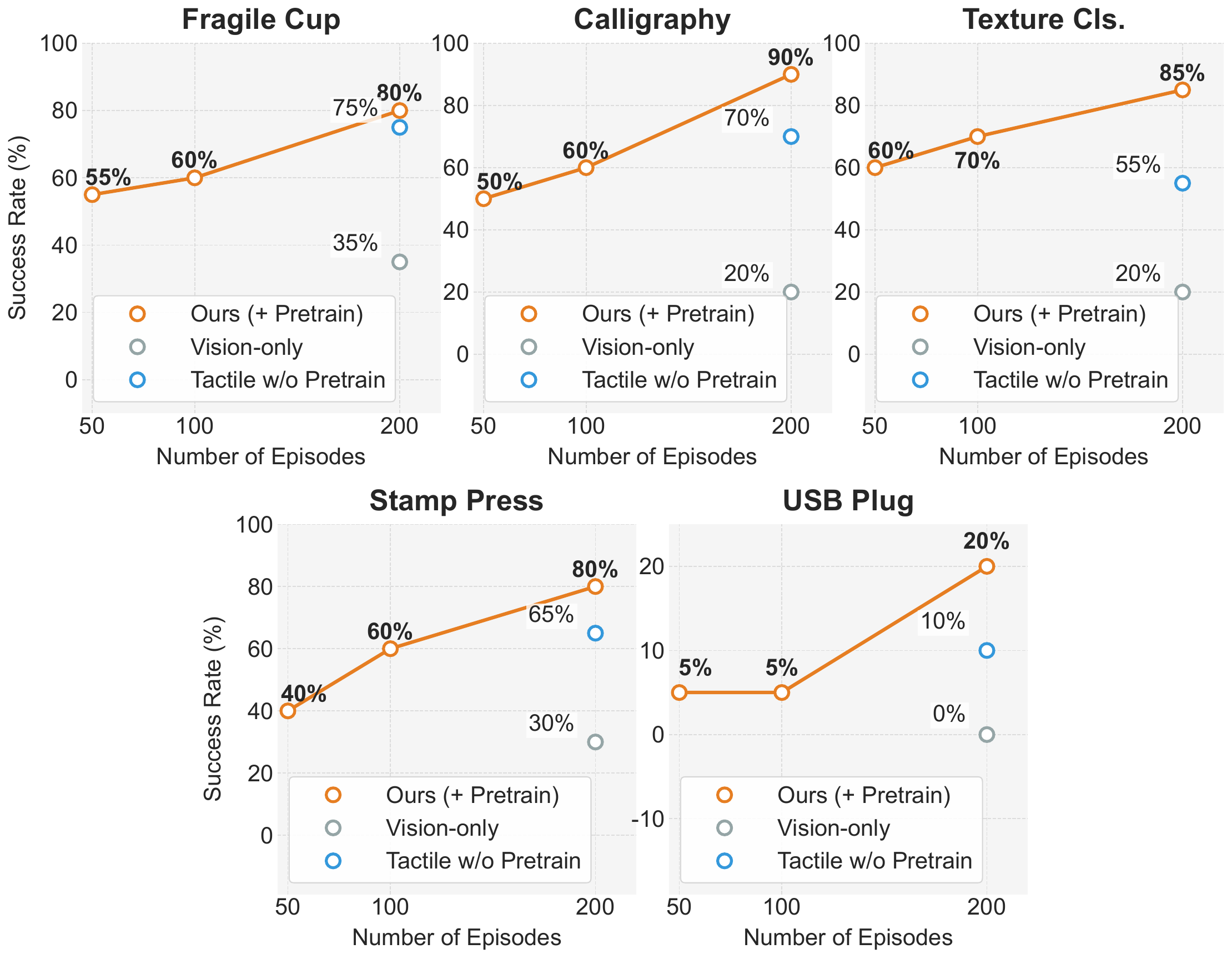}
      \caption{\textbf{Ablation across training episodes.} Our pretrained policy, even with small data, outperforms the vision-only policy trained with large amounts of data. Moreover, it approaches the performance of a large-data tactile-only policy.}
      \label{fig:dataefficiency}
      \vspace{-10pt}
   \end{figure}

\textbf{Pretraining demonstrates substantial benefits under data-scarce conditions (Q3).} As shown in \cref{fig:dataefficiency}, ACT policy with pretrained encoder achieves superior performance with only 50 task-specific demonstrations—55\% for cup manipulation, 50\% for calligraphy and 60\% for texture classification—compared to the visual-only baseline (35\%, 20\%, and 20\%, respectively). With 100 episodes, its performance in cup and stamp tasks (60\% each) nearly matches the non-pretrained tactile model (70\% and 65\%). We further analyze performance discrepancies across task types. 
Perception-centric tasks like texture classification benefit from the generalized features of the pre-training, while action-oriented tasks such as cup manipulation and calligraphy require not only robust features but also sufficient task-specific demonstrations to learn effective action strategies.

\begin{table}[t]
\caption{\textbf{Cross-sensor generalization results.}
Visuo-tactile pretraining helps generalization to other visuo-tactile sensor setups. ID: in-domain. OOD: out-of-domain.}
\label{tab:crosssensor}
\centering
\small 
\begin{tabular}{llcccc}
\toprule
\textbf{Task} & \textbf{Phase} & \multicolumn{2}{c}{\textbf{\makecell{Ours (+ Tactile\\ w/o Pretrain)}}} & \multicolumn{2}{c}{\textbf{\makecell{Ours \\(+ Pretrain)}}} \\
\cmidrule(lr){3-4} \cmidrule(lr){5-6}
 &  & \makecell{ID\\ Sensor} & \makecell{OOD\\ Sensor} & \makecell{ID\\ Sensor} & \lightgray\makecell{OOD\\ Sensor} \\
\midrule
\multirow{3}{*}{\makecell[l]{Texture\\Cls.}} 
 & Grasp       & 75\% & \textbf{80\%} & 85\% & \lightgray\textbf{85\%} \\
 & Classify    & 55\% & \textbf{40\%} & 85\% & \lightgray\textbf{75\%} \\
 &Whole Task  & 55\% & \textbf{40\%} & 85\% & \lightgray\textbf{75\%} \\
\midrule
\multirow{3}{*}{\makecell[l]{Fragile\\Cup}} 
 & Pick        & 80\% & \textbf{70\%} & 80\% & \lightgray\textbf{80\%} \\
 & Land        & 75\% & \textbf{10\%} & 80\% & \lightgray\textbf{80\%} \\
 & Whole Task  & 75\% & \textbf{10\%} & 80\% & \lightgray\textbf{80\%} \\
\bottomrule
\end{tabular}
\vspace{-10pt} 
\end{table}

\textbf{Pretraining on \methodname dataset provides robustness against variations among visuo-tactile sensors (Q3).} 
While our main experiments are conducted in an in-domain setting, where training and evaluation use different sensor batches with high consistency, we also evaluate a more challenging out-of-domain scenario by introducing substantial variations across visuo-tactile sensors.
As shown in~\Cref{tab:crosssensor}, the pretrained model generalizes well across two distinct sensors, which differ in marker angle (0° vs. 45°) and RGB lighting environment (side vs. bottom). The in-domain and out-of-domain performances remain close on the texture classification task (0.85 vs. 0.75) and identical on the fragile cup task (both 0.8). In contrast, the non-pretrained counterpart performs poorly, particularly in phases requiring fine-grained tactile perception: a drop from 80\% to 40\% in the tactile-based classification phase, and a catastrophic drop from 70\% to 10\% in the landing phase (which relies on detecting contact with the table before releasing). These results validate the effectiveness of tactile pretraining for cross-sensor generalization.

\section{Conclusion}
We present~\methodname, a human-centric and robot-free data collection system with in-situ visuo-tactile feedback and recording. 
An in-situ and modular gripper with visuo-tactile sensors enables rapid adaptation, and a large-scale high-precision dataset is collected to support contact-rich manipulation policy learning. 
Experimental results demonstrate that the proposed system outperforms existing methods in multiple aspects, including data collection efficiency, control accuracy, and human-machine interaction experience. Through policy validation, the effectiveness of the collected data and the importance of visuo-tactile pretraining are further confirmed.

\textbf{Limitation and future work.} While \methodname has demonstrated efficacy across a range of challenging tasks, a few limitations remain. 
To eliminate dependency on external base stations for submillimeter-level localization, we will develop high-precision visual algorithms for collecting data in the wild.
We also plan to extend our system and dataset to dexterous hands and bimanual long-horizon tasks, facilitating study on finer-grained and more complex scenarios.

\section{Acknowledgments}
We gratefully acknowledge Huijie Wang for helping to develop the demonstration page, and Zherui Qiu for assisting with the supervision of the user study. We also thank Yixuan Pan, Qingwen Bu, Zhuoheng Li, Jisong Cai, Yuxiang Lu, and Ningbin Zhang for their valuable insights and constructive discussions. Our appreciation also goes to Yingxin Wang, Zhirui Zhang, Xinyu Yang, Fengjie Shen, Taoyuan Huang, and Lekai Chen for their assistance during the data collection. Finally, we extend our sincere gratitude to JAKA for their generous support in the collection of our dataset. This work is in part supported by the JC STEM Lab of Autonomous Intelligent Systems funded by The Hong Kong Jockey Club Charities Trust. This work was also partially supported by the NSFC Grant No.52505029, and the STCSM Grant No.25ZR1401191 and No. 24511103400.

{
\small
\bibliographystyle{IEEEtran}
\bibliography{bibliography_short, refs}  %

\begin{thebibliography}{10}
\providecommand{\url}[1]{#1}
\csname url@samestyle\endcsname
\providecommand{\newblock}{\relax}
\providecommand{\bibinfo}[2]{#2}
\providecommand{\BIBentrySTDinterwordspacing}{\spaceskip=0pt\relax}
\providecommand{\BIBentryALTinterwordstretchfactor}{4}
\providecommand{\BIBentryALTinterwordspacing}{\spaceskip=\fontdimen2\font plus
\BIBentryALTinterwordstretchfactor\fontdimen3\font minus \fontdimen4\font\relax}
\providecommand{\BIBforeignlanguage}[2]{{%
\expandafter\ifx\csname l@#1\endcsname\relax
\typeout{** WARNING: IEEEtran.bst: No hyphenation pattern has been}%
\typeout{** loaded for the language `#1'. Using the pattern for}%
\typeout{** the default language instead.}%
\else
\language=\csname l@#1\endcsname
\fi
#2}}
\providecommand{\BIBdecl}{\relax}
\BIBdecl

\bibitem{connor1992neural}
C.~E. Connor and K.~O. Johnson, ``Neural coding of tactile texture: comparison of spatial and temporal mechanisms for roughness perception,'' \emph{Journal of Neuroscience}, 1992.

\bibitem{xu2023visual}
W.~Xu, Z.~Yu, H.~Xue, R.~Ye, S.~Yao, and C.~Lu, ``Visual-tactile sensing for in-hand object reconstruction,'' in \emph{CVPR}, 2023.

\bibitem{tiest2009cues}
W.~M.~B. Tiest and A.~M. Kappers, ``Cues for haptic perception of compliance,'' \emph{IEEE Trans. on Haptics}, 2009.

\bibitem{ma2019dense}
D.~Ma, E.~Donlon, S.~Dong, and A.~Rodriguez, ``Dense tactile force estimation using gelslim and inverse fem,'' in \emph{ICRA}, 2019.

\bibitem{yang2026riseselfimprovingrobotpolicy}
J.~Yang, K.~Lin, J.~Li, W.~Zhang, T.~Lin, L.~Wu, Z.~Su, H.~Zhao, Y.-Q. Zhang, L.~Chen \emph{et~al.}, ``{RISE}: Self-improving robot policy with compositional world model,'' \emph{arXiv preprint arXiv:2602.11075}, 2026.

\bibitem{shi2026egohumanoidunlockinginthewildlocomanipulation}
M.~Shi, S.~Peng, J.~Chen, H.~Jiang, Y.~Li, D.~Huang, P.~Luo \emph{et~al.}, ``{EgoHumanoid}: Unlocking in-the-wild loco-manipulation with robot-free egocentric demonstration,'' \emph{arXiv preprint arXiv:2602.10106}, 2026.

\bibitem{shi2025diversity}
M.~Shi, L.~Chen, J.~Chen, Y.~Lu, C.~Liu, G.~Ren, P.~Luo, D.~Huang, M.~Yao, and H.~Li, ``Is diversity all you need for scalable robotic manipulation?'' \emph{TRO}, 2026.

\bibitem{jiang2025wholebodyvla}
H.~Jiang, J.~Chen, Q.~Bu, L.~Chen, M.~Shi, Y.~Zhang, D.~Li, C.~Suo, C.~Wang, Z.~Peng \emph{et~al.}, ``{WholeBodyVLA}: Towards unified latent vla for whole-body loco-manipulation control,'' in \emph{ICLR}, 2026.

\bibitem{padalkar2023open}
A.~Padalkar, A.~Pooley, A.~Jain, A.~Bewley, A.~Herzog, A.~Irpan, A.~Khazatsky, A.~Rai, A.~Singh, A.~Brohan \emph{et~al.}, ``{Open X-Embodiment}: Robotic learning datasets and {RT-X} models,'' in \emph{ICRA}, 2024.

\bibitem{bu2025agibot}
Q.~Bu, J.~Cai, L.~Chen, X.~Cui, Y.~Ding, S.~Feng, S.~Gao, X.~He, X.~Huang \emph{et~al.}, ``{AgiBot World Colosseo}: A large-scale manipulation platform for scalable and intelligent embodied systems,'' \emph{IROS}, 2025.

\bibitem{chen2025intelligent}
L.~Chen, C.~Sima, K.~Chitta, A.~Loquercio, P.~Luo, Y.~Ma, and H.~Li, ``Intelligent robot manipulation requires self-directed learning,'' \emph{Authorea Preprints}, 2025.

\bibitem{lin2025learning}
T.~Lin, Y.~Zhang, Q.~Li, H.~Qi, B.~Yi, S.~Levine, and J.~Malik, ``Learning visuotactile skills with two multifingered hands,'' in \emph{ICRA}, 2025.

\bibitem{huang20243d}
B.~Huang, Y.~Wang, X.~Yang, Y.~Luo, and Y.~Li, ``{3D-ViTac}: Learning fine-grained manipulation with visuo-tactile sensing,'' in \emph{CoRL}, 2024.

\bibitem{xue2025reactive}
H.~Xue, J.~Ren, W.~Chen, G.~Zhang, Y.~Fang, G.~Gu, H.~Xu, and C.~Lu, ``{Reactive Diffusion Policy}: Slow-fast visual-tactile policy learning for contact-rich manipulation,'' in \emph{RSS}, 2025.

\bibitem{fu2024mobile}
Z.~Fu, T.~Z. Zhao, and C.~Finn, ``{Mobile ALOHA}: Learning bimanual mobile manipulation with low-cost whole-body teleoperation,'' in \emph{CoRL}, 2024.

\bibitem{chi2024universal}
C.~Chi, Z.~Xu, C.~Pan, E.~Cousineau, B.~Burchfiel, S.~Feng, R.~Tedrake, and S.~Song, ``{Universal Manipulation Interface}: In-the-wild robot teaching without in-the-wild robots,'' in \emph{RSS}, 2024.

\bibitem{zhaxizhuoma2025fastumi}
Z.~Zhaxizhuoma, K.~Liu, C.~Guan, Z.~Jia, Z.~Wu, X.~Liu, T.~Wang, S.~Liang \emph{et~al.}, ``{Fast-UMI}: A scalable and hardware-independent universal manipulation interface with dataset,'' in \emph{CoRL}, 2025.

\bibitem{dou2024tactile}
Y.~Dou, F.~Yang, Y.~Liu, A.~Loquercio, and A.~Owens, ``Tactile-augmented radiance fields,'' in \emph{CVPR}, 2024.

\bibitem{cheng2025touch100k}
N.~Cheng, J.~Xu, C.~Guan, J.~Gao, W.~Wang, Y.~Li, F.~Meng, J.~Zhou, B.~Fang \emph{et~al.}, ``Touch100k: A large-scale touch-language-vision dataset for touch-centric multimodal representation,'' \emph{Information Fusion}, 2025.

\bibitem{tao2011situ}
F.~Tao and M.~Salmeron, ``In situ studies of chemistry and structure of materials in reactive environments,'' \emph{Science}, 2011.

\bibitem{zheng2024high}
W.~Zheng, P.~Chai, J.~Zhu, and K.~Zhang, ``High-resolution in situ structures of mammalian respiratory supercomplexes,'' \emph{Nature}, 2024.

\bibitem{chen2024arcap}
S.~Chen, C.~Wang, K.~Nguyen, L.~Fei-Fei, and C.~K. Liu, ``{ARCap}: Collecting high-quality human demonstrations for robot learning with augmented reality feedback,'' in \emph{ICRA}, 2025.

\bibitem{wang2024dexcap}
C.~Wang, H.~Shi, W.~Wang, R.~Zhang, L.~Fei-Fei, and C.~K. Liu, ``{DexCap}: Scalable and portable mocap data collection system for dexterous manipulation,'' in \emph{RSS}, 2024.

\bibitem{bunny-visionpro}
R.~Ding, Y.~Qin, J.~Zhu, C.~Jia, S.~Yang, R.~Yang, X.~Qi, and X.~Wang, ``{Bunny-VisionPro}: Real-time bimanual dexterous teleoperation for imitation learning,'' \emph{arXiv preprint arXiv:2407.03162}, 2024.

\bibitem{wu2024gello}
P.~Wu, Y.~Shentu, Z.~Yi, X.~Lin, and P.~Abbeel, ``Gello: A general, low-cost, and intuitive teleoperation framework for robot manipulators,'' in \emph{IROS}, 2024.

\bibitem{buamanee2024bi}
T.~Buamanee, M.~Kobayashi, Y.~Uranishi, and H.~Takemura, ``{Bi-ACT}: Bilateral control-based imitation learning via action chunking with transformer,'' in \emph{AIM}, 2024.

\bibitem{liu2025vitamin}
F.~Liu, C.~Li, Y.~Qin, A.~Shaw, J.~Xu, P.~Abbeel, and R.~Chen, ``{ViTaMIn}: Learning contact-rich tasks through robot-free visuo-tactile manipulation interface,'' \emph{arXiv preprint arXiv:2504.06156}, 2025.

\bibitem{pan2025agility}
Y.~Pan, R.~Qiao, L.~Chen, K.~Chitta, L.~Pan, H.~Mai, Q.~Bu, H.~Zhao, C.~Zheng, P.~Luo, and H.~Li, ``{Agility Meets Stability}: Versatile humanoid control with heterogeneous data,'' in \emph{ICRA}, 2025.

\bibitem{zhao2023act}
T.~Z. Zhao, V.~Kumar, S.~Levine, and C.~Finn, ``Learning fine-grained bimanual manipulation with low-cost hardware,'' in \emph{RSS}, 2023.

\bibitem{chi2023diffusion}
C.~Chi, Z.~Xu, S.~Feng, E.~Cousineau, Y.~Du \emph{et~al.}, ``Diffusion policy: Visuomotor policy learning via action diffusion,'' \emph{IJRR}, 2024.

\bibitem{brygo2014humanoid}
A.~Brygo, I.~Sarakoglou, N.~Garcia-Hernandez, and N.~Tsagarakis, ``Humanoid robot teleoperation with vibrotactile based balancing feedback,'' in \emph{EuroHaptics}, 2014.

\bibitem{Huang_2025}
H.-J. Huang, M.~Kaess, and W.~Yuan, ``{NormalFlow}: Fast, robust, and accurate contact-based object 6dof pose tracking with vision-based tactile sensors,'' \emph{RAL}, 2025.

\bibitem{DBLP:conf/rss/KerrHWHICG23}
J.~Kerr, H.~Huang, A.~Wilcox, R.~Hoque, J.~Ichnowski, R.~Calandra, and K.~Goldberg, ``Self-supervised visuo-tactile pretraining to locate and follow garment features,'' in \emph{RSS}, 2023.

\bibitem{fu2024touchvisionlanguagedataset}
L.~Fu \emph{et~al.}, ``A touch, vision, and language dataset for multimodal alignment,'' \emph{arXiv preprint arXiv:2402.13232}, 2024.

\bibitem{gao2023objectfolder}
R.~Gao, Y.~Dou, H.~Li, T.~Agarwal, J.~Bohg, Y.~Li, L.~Fei-Fei, and J.~Wu, ``The objectfolder benchmark: Multisensory learning with neural and real objects,'' in \emph{CVPR}, 2023.

\bibitem{10710144}
T.~Li, Y.~Yan, C.~Yu, J.~An, Y.~Wang, X.~Zhu, and G.~Chen, ``Vtg: A visual-tactile dataset for three-finger grasp,'' \emph{RAL}, 2024.

\bibitem{liu2025vtdexmanip}
Q.~Liu, Y.~Cui, Z.~Sun, G.~Li, J.~Chen, and Q.~Ye, ``{VTD}exmanip: A dataset and benchmark for visual-tactile pretraining and dexterous manipulation with reinforcement learning,'' in \emph{ICLR}, 2025.

\bibitem{zhu2025touchwildlearningfinegrained}
X.~Zhu, B.~Huang, and Y.~Li, ``Touch in the wild: Learning fine-grained manipulation with a portable visuo-tactile gripper,'' in \emph{NeurIPS}, 2025.

\bibitem{DBLP:conf/icira/RenZG23}
J.~Ren, J.~Zou, and G.~Gu, ``{MC-Tac}: Modular camera-based tactile sensor for robot gripper,'' in \emph{ICIRA}, 2023.

\bibitem{zoeller2020systematic}
A.~C. Zoeller and K.~Drewing, ``A systematic comparison of perceptual performance in softness discrimination with different fingers,'' \emph{Attention, Perception, \& Psychophysics}, 2020.

\bibitem{Manceron_IKPy}
P.~Manceron, ``{IKPy},'' \url{https://github.com/Phylliade/ikpy}, 2016.

\bibitem{george2024visuo}
A.~George, S.~Gano, P.~Katragadda, and A.~B. Farimani, ``Visuo-tactile pretraining for cable plugging,'' in \emph{ICRA}, 2024.

\bibitem{radford2021learning}
A.~Radford, J.~W. Kim, C.~Hallacy, A.~Ramesh, G.~Goh, S.~Agarwal, G.~Sastry, A.~Askell, P.~Mishkin, J.~Clark \emph{et~al.}, ``Learning transferable visual models from natural language supervision,'' in \emph{ICML}, 2021.

\end{thebibliography}
}

\clearpage
\begin{appendices}
\setcounter{section}{0}

\renewcommand\thesection{\Alph{section}}      %
\renewcommand\thesubsection{\thesection.\arabic{subsection}} 

\section{Hardware Design}
\label{app:hardware}

\textcolor{gray}{\textit{| Supplement to \cref{sec:method-hardware} in the Main paper.}}

\textbf{Hardware implementation details.}
The sensor is rigidly connected to the gripper base via a primary threaded hole, bearing the main mechanical load. On the opposite side, a protruding feature fits precisely into a corresponding groove on the gripper base, enabling straightforward plug-in alignment and positioning. Additional constraint is provided by auxiliary screws on the rear side to prevent damage from vibration or impact, further enhancing stability.

For visual perception, we use a fisheye camera equipped with a $180^\circ$ field-of-view lens, capturing video at 30 frames per second with a resolution of $640 \times 480$ pixels. The tactile sensor is integrated with a camera operating at 30 FPS with a resolution of $640 \times 480$ pixels.
Designed with custom 3D-printed parts and commercially available standard components, the system achieves a lightweight ($157.5 g$) and compact ($145 \times 85 \times 106 {mm}^3$) form factor, ensuring portability and reproducibility. 

\begin{figure}[ht]
    \centering
    \includegraphics[width=\columnwidth]{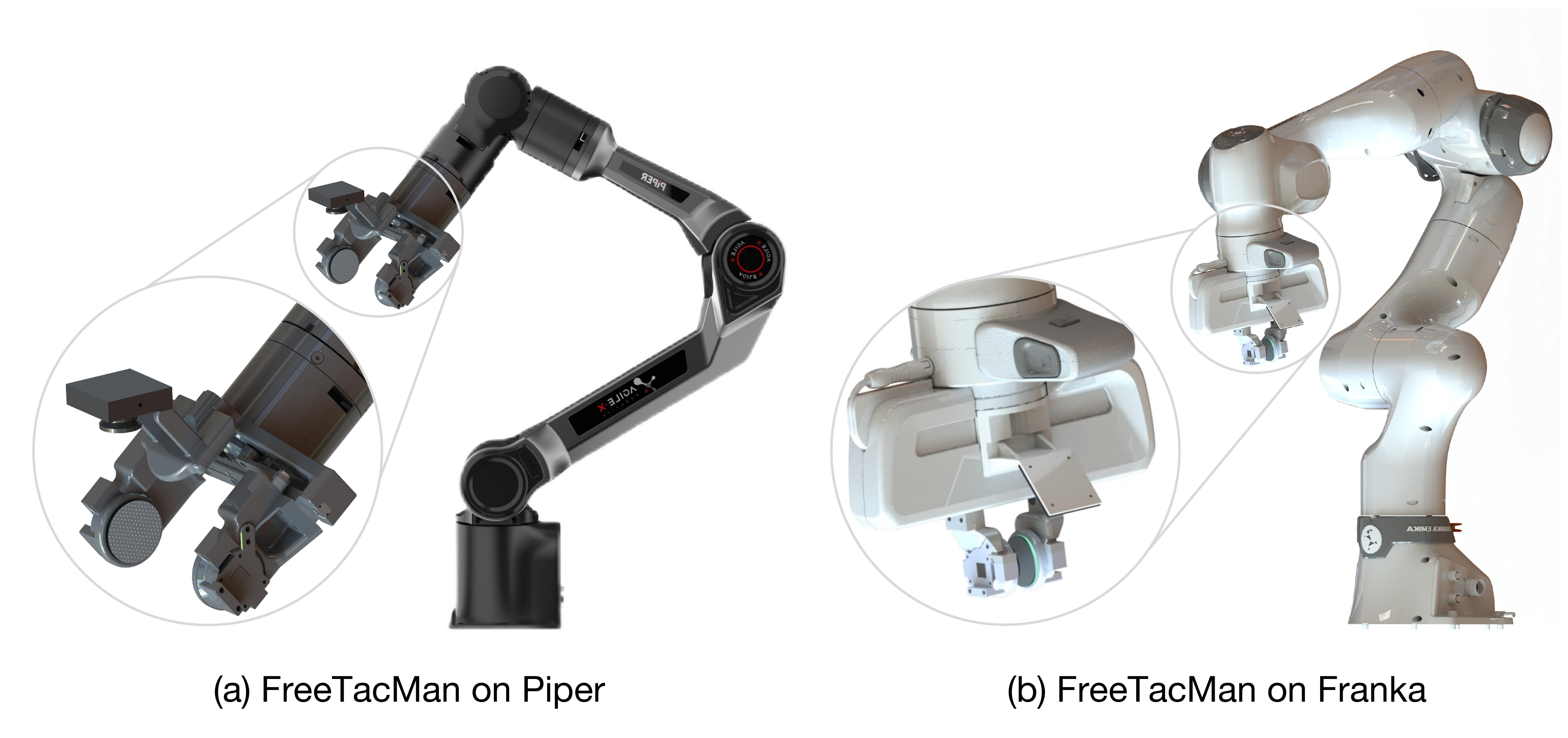}
    \caption{\textbf{Detailed mounting interface for different robot arms.} (a) and (b) illustrate the integration of \methodname with the PIPER and Franka robotic arms, respectively.
    The zoomed-in views highlight the connector regions and camera mounting positions, which are carefully aligned to maintain a consistent viewpoint between the data collection and execution systems.
    }
    \label{fig:render}
\end{figure}

\textbf{Seamless integration into robotic platforms.}
\cref{fig:render} illustrates the universal gripper interface with quick-swap mounts compatible with both the Piper and Franka arms, as well as the camera scaffold designed for precise alignment with the wrist-mounted camera to ensure consistent perspective. These components demonstrate the plug-and-play modularity of \methodname, enabling seamless integration across diverse robotic platforms without requiring hardware-specific modifications. 

\section{Data Processing}

\textcolor{gray}{\textit{| Supplement to \cref{sec:method-data} in the Main paper.}}

\begin{figure}[t]
    \centering
    \includegraphics[width=\columnwidth]{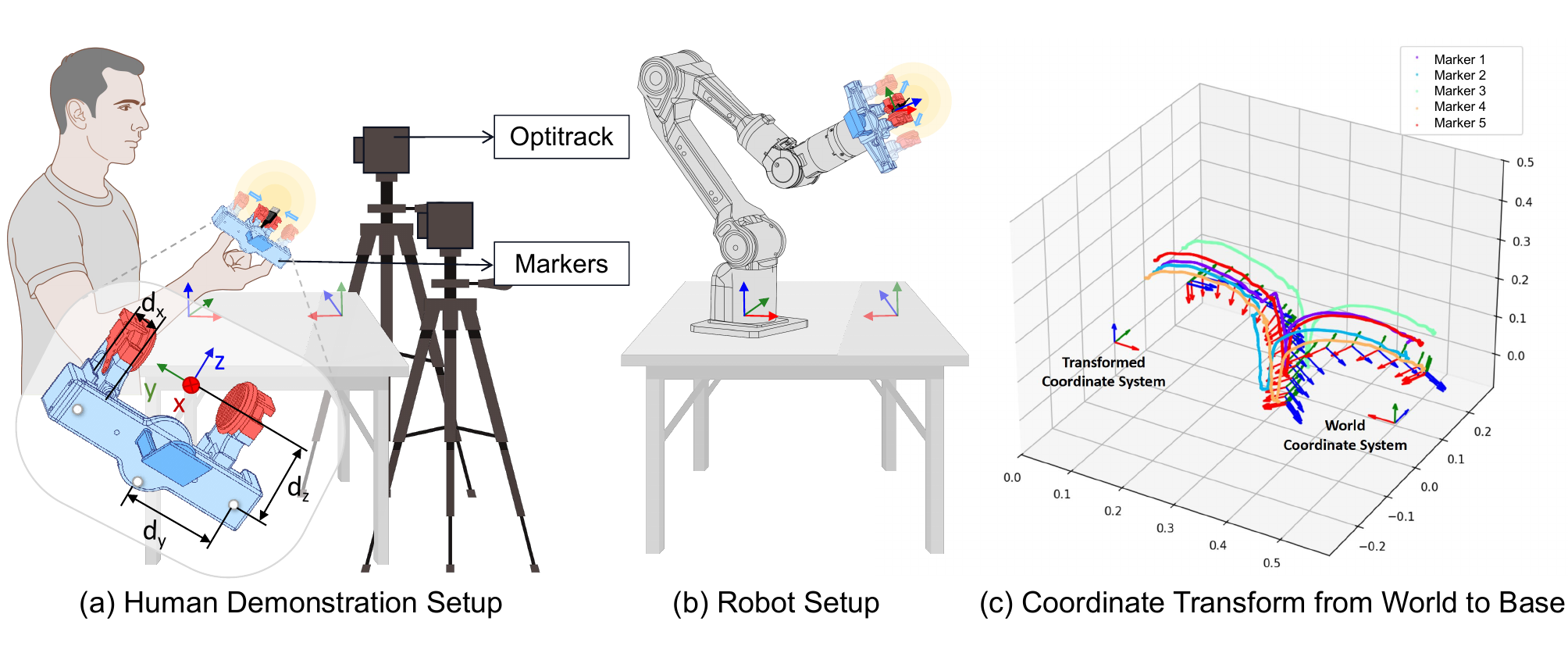}
    \caption{\textbf{System setup and coordinate transformation for human demonstration and robot execution.} (a) Human demonstration setup showing the OptiTrack-defined world frame, the robot base frame, and the local frame on the data collection device. (b) Robot setup with the base at the origin and a local frame consistent with the demonstration. (c) Coordinate transformation example for texture classification task, visualizing five marker trajectories and the local frame in 3D space.}
    \label{fig:setup}
\end{figure}

Fig. \ref{fig:setup} elaborates on coordinate transformation for demonstration-to-execution consistency.
During human demonstration, the OptiTrack system defines a global coordinate frame, which differs from the robot’s base frame used during inference. To ensure consistency, the positions of the five markers are transformed—via rotation and translation—into the robot’s base frame for accurate task execution.
To obtain the robot end-effector pose—necessary for subsequent joint angle computation—a local coordinate frame is established with the Tool Center Point (TCP) as its origin. All coordinate frames follow the right-hand Cartesian convention. The position of TCP is determined using three markers mounted on the top plate. Among these, the two with the greatest distance define the direction of the $dy$ axis, while the $dx$ axis points from the third marker to the midpoint of the other two. The magnitude of $dy$ is set to half the distance between the two farthest markers. The $dx$ and $dz$ offsets, defined by hardware dimensions, represent the TCP’s position relative to the two front-facing markers.
During task execution, this local coordinate frame remains aligned with the one defined during data collection.
Fig.~\ref{fig:setup}(c) illustrates an example of coordinate transformation in a texture classification task, visualizing the trajectories of five markers and the local frame in 3D space.

\section{Training Details}
\label{app:pretrain}
\textcolor{gray}{\textit{| Supplement to \cref{sec:method-policy} in the Main paper.}}

\textbf{Problem formulation.}
We define a visuo-tactile manipulation policy for the robot as a mapping $\pi: \mathcal{O} \rightarrow \mathcal{A}$, 
where the observation space \(\mathcal{O}\) consists of three modalities: visual observation $\mathbf{o}_t^v\in\mathcal{O}^v\subset\mathbb{R}^{H\times W\times 3}$, tactile observation $\mathbf{o}_t^t\in\mathcal{O}^t\subset\mathbb{R}^{H\times W\times 3}$, and robot proprioception $\mathbf{o}_t^r\in\mathcal{O}^r\subset\mathbb{R}^{n_s}$ as 
$\mathbf{o} = \left( \mathbf{o}_t^v, \mathbf{o}_t^t, \mathbf{o}_t^r \right)$. The action space $\mathbf{a}\in\mathcal{A}\subset\mathbb{R}^7$ is defined as 6-DoF arm joint position and 1-DoF gripper position. The policy $\pi$ is learned via imitation learning to map these observations to corresponding actions.

\begin{table}[b]
    \centering
    \caption{Hyperparameters for tactile pretraining.}
    \label{tab:hyper_pretrain}
    \footnotesize
    \begin{tabular}{l c}
        \toprule
        \textbf{Hyperparameters} & \textbf{Value}   \\
        \midrule
         Visual backbone & ResNet-18 (Frozen)       \\
         Tactile backbone & ResNet-18 (finetuned)   \\
         Projection dimension & $256$                 \\
         Learning rate & $1e-4$                       \\
         Batch size & 128                              \\
         \bottomrule
          
    \end{tabular}
\end{table}

\textbf{Details on tactile pretraining.}
We pretrain the tactile encoder $f_t$ and projection head $g_v$, $g_t$ using a CLIP-style contrastive loss with multi-positive sampling.  
Optimization is performed using AdamW with cosine-annealed learning rates, updating only the tactile encoder $f_t$, visual $g_v$, and tactile projection $g_t$ while keeping visual encoder $f_v$ fixed. We train on the sequence of time-aligned visuo-tactile frames, drawing primary positives from the same timestep and secondary positives from the next frame (with wrap-around). Negatives are sampled from a memory bank of size $4096$. Table \ref{tab:hyper_pretrain} shows the hyperparameters of pretraining.

\textbf{Details on policy learning.}
We adopt the action chunking transformer (ACT)~\cite{zhao2023act} architecture to learn visuo-tactile manipulation policies. Both visual and tactile images are \(\,640\times480\) pixels and fed as stacked inputs to the ACT backbones. At each timestep, the model outputs an action chunk of length $\mathcal{T}$, where each action in the chunk specifies a 6-DOF joint action for robot arm plus a 1-DOF gripper action.
To integrate our visuo-tactile pipeline, we add our pretrained tactile encoder \(f_t\) to extract tactile features. The tactile features are concatenated with visual features and input into the transformer encoder of ACT. \Cref{tab:hyper_policy} shows the hyperparameters of policy learning.

\begin{table}[t]
    \centering
    \caption{Hyperparameters for policy learning.}
    \label{tab:hyper_policy}
    \footnotesize
    \begin{tabular}{l c}
        \toprule
        \textbf{Hyperparameters} & \textbf{Value}   \\
        \midrule
         Visual RGB resolution & \(640\times480\)        \\
         Tactile RGB resolution & \(640\times480\)        \\
         Chunk size & $48$                         \\
         Hidden dimension & 512                     \\
         Visual backbone & ResNet-18      \\
         Tactile backbone & ResNet-18 (Pretrained)   \\
         Learning rate & $4e-5$                       \\
         Weight decay & $1e-4$                          \\
         Batch size & $64$                              \\
         KL weight & $10$                                \\
         \bottomrule
    \end{tabular}
\end{table}

\section{Experimental Setup}
\textcolor{gray}{\textit{| Supplement to \cref{sec:exp-setup} in the Main paper.}}
\label{experiment setup}
\begin{figure}[b]
    \centering
    \includegraphics[width=\columnwidth]{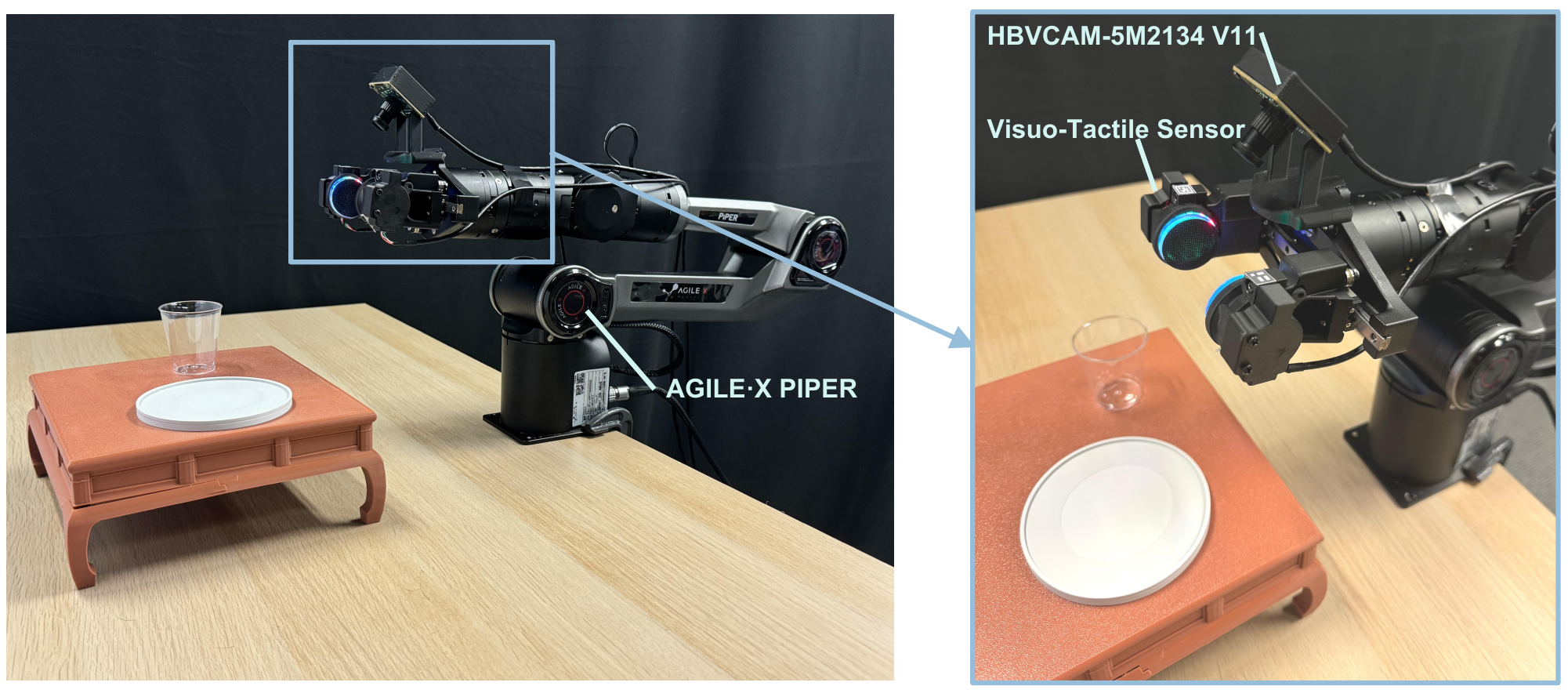}
    \caption{\textbf{Experiment setup for policy deployment.} The sensing system, consisting of a fisheye camera and two tactile sensors, is mounted on the end effector in the same configuration used during data collection.}
    \label{fig:realrobot}
\end{figure}

\textbf{Task specification.}
We present a comprehensive description of 
each task below.
\begin{itemize} [leftmargin=*,itemsep=0pt,topsep=0pt]
    \item \textbf{Fragile cup.} In this task, the robot is required to grasp a fragile cup from a rack and place it into a tray on the rack. A trial is considered successful if the cup is stably placed within the tray without any damage or visible deformation to the cup. This task is designed to evaluate the role of tactile feedback in delicate object handling.
    \item \textbf{USB plug.} The robot must insert a pre-grasped USB into a specified socket on a power strip for this task. A success is recorded if the USB is inserted to a sufficient depth such that it remains securely connected. Tactile information serves as an additional signal to assist with fine alignment.
    \item \textbf{Texture classification.} This task involves recognizing and sorting two types of cylindrical objects, each covered with distinct industrial-grade surface textures. The robot must grasp an object, identify its texture through visuo-tactile perception, and place it into the correct bin corresponding to the texture type. Tactile sensing provides direct information for surface material recognition.
    \item \textbf{Stamp press.} In the stamping task, the robot needs to use a pre-grasped, spring-loaded stamp to produce a clear and complete imprint onto a sheet of paper laid flat on the table. The task is deemed successful if a clear and complete imprint of the stamp is produced on the paper surface. The stamping task leverages tactile feedback to control contact force during imprinting.
    \item \textbf{Calligraphy.} For the calligraphy brush writing task, the robot must use a pre-grasped calligraphy brush to trace the digit "5" following a guide labeled "202" on a piece of cloth. A success is determined if the resulting written pattern can be visually recognized and interpreted as the number "5". During this task, tactile information assists the robot in both estimating the in-hand brush pose for fine motion control and ensuring the brush-lifting action between the first and the second stroke.
\end{itemize}

\textbf{Implementation details.}
\label{app:impl}
We conduct all experiments on a PIPER 6-DOF light-weight robotic arm (AGILE·X Robotics) equipped with our \methodname mounted at the wrist, as shown in \cref{fig:realrobot}. The robot is connected to a workstation running Ubuntu 20.04 and ROS Noetic on an NVIDIA RTX 4090 GPU. 
RGB frames are captured at 30 Hz by HBVCAM fisheye camera and published via ROS topics, and tactile frames from our camera-based tactile sensor are sampled at 30 Hz.  Our imitation policy performs inference on the GPU, with policy latency averaging under 20 ms per cycle. 

\textbf{Evaluation metrics in user study.}
We provide detailed explanations of the evaluation metrics used for user experience evaluation. 
\begin{itemize}[leftmargin=*,itemsep=0pt,topsep=0pt]
    \item \textbf{Comfort and ease of collection:} After completing five tasks with each of the three systems, participants rated how physically comfortable it was to perform the demonstrations on a scale from 1 (very uncomfortable) to 5 (very comfortable). In the same survey, they scored the ease of collection from 1 (very difficult) to 5 (very easy). 
    \item \textbf{Ease of entry:} After using all three systems, participants ranked them in order of entry difficulty.  We convert these rankings into a normalized score where a lower average rank corresponds to a higher ease-of-entry value.
    \item \textbf{Control accuracy:} Volunteers ordered the three systems by how precisely they could control the gripper during demonstrations. These ranks are normalized so that a lower mean rank indicates higher control fidelity.
    \item \textbf{Stability and damage prevention:} During each demonstration, we logged the number of times the prop slipped from the gripper and the number of times the prop was visibly deformed or damaged.
\end{itemize}
All metric values are normalized, with higher scores indicating better performance, as visualized by the radar chart shown in Fig.~\ref{fig:userstudy}(d) in the main paper.

\section{Additional Experiments}
\label{app:exp}

\textbf{Evaluation on user study.}
Table \ref{tab:user_study} reports completion rate, task duration, slip count, damage count and CPUT for each method across five contact-rich tasks.
\begin{table*}[t]
  \centering
  \footnotesize
  \caption{
    \textbf{Detailed results of user study.} \methodname outperforms existing data collection methods across five manipulation tasks. 
    \textbf{Completion Rate} is the fraction of successful demonstrations;
    \textbf{Time} is the average collection duration;
    \textbf{Slip Count} and \textbf{Damage Count} record object slips and damages during collection;
    \textbf{Performance} is an overall metric combining efficiency and reliability.
  }
  \setlength{\tabcolsep}{2pt}
  \begin{tabular*}{\textwidth}{@{\extracolsep{\fill}} l l c c c c c }
    \toprule
    \textbf{Task} & \textbf{Method} & \textbf{Complet. Rate} & \textbf{Task Duration (s)} 
      & \textbf{Slip Count} & \textbf{Damage Count} & \textbf{CPUT Score} \\
    \midrule
    \multirow{3}{*}{Texture Cls.}
      & ALOHA   & 0.8611 & 11.41 &  5 &  0 & 0.0495 \\
      & UMI     & 1.0000 &  3.27 &  0 &  0 & 0.3060 \\
      & Ours    & 1.0000 &  3.15 &  0 &  0 & 0.3171 \\
    \midrule
    \multirow{3}{*}{Fragile Cup}
      & ALOHA   & 0.5274 & 11.19 &  2 & 14 & 0.0471 \\
      & UMI                     & 1.0000 &  4.19 &  0 &  0 & 0.2386 \\
      & Ours                        & 1.0000 &  3.50 &  0 &  0 & 0.2854 \\
    \midrule
    \multirow{3}{*}{USB Plug}
      & ALOHA   & 0.6108 & 12.63 &  9 &  0 & 0.0483 \\
      & UMI                     & 0.2220 &  5.55 & 27 &  0 & 0.0400 \\
      & Ours                        & 0.9722 &  4.24 &  2 &  0 & 0.2292 \\
    \midrule
    \multirow{3}{*}{Calligraphy}
      & ALOHA   & 0.5553 & 14.89 &  2 &  0 & 0.0373 \\
      & UMI                     & 1.0000 &  5.03 &  2 &  0 & 0.1989 \\
      & Ours                        & 1.0000 &  4.17 &  0 &  0 & 0.2401 \\
    \midrule
    \multirow{3}{*}{Stamp Press}
      & ALOHA   & 0.7775 &  8.09 &  3 &  2 & 0.0962 \\
      & UMI                     & 0.5553 &  3.91 & 26 &  0 & 0.1421 \\
      & Ours                        & 1.0000 &  2.98 &  1 &  0 & 0.3356 \\
    \bottomrule
  \end{tabular*}
  \label{tab:user_study}
\end{table*}
\begin{itemize}[leftmargin=*,itemsep=0pt,topsep=0pt]
    \item \textbf{Completion rate:} Across all tasks, \methodname achieves the highest completion rates, reaching 100\% (Texture classification, Fragile cup, Stamp press, Calligraphy) or near-perfect rates (USB plug). In contrast, ALOHA's rates range from 52.74\% (Fragile cup) to 86.11\% (Texture classification), and UMI's range from 22.20\% (USB plug) to 100\% (Texture classification, Fragile cup, Calligraphy). This demonstrates that real-time tactile feedback enhances the human operator's ability to execute and complete precise, contact-rich tasks.
    \item \textbf{Task duration:} \methodname demonstrates the shortest task durations, averaging between 2.98s (Stamp press) and 4.24s (USB plug). UMI's times range from 3.27s to 5.55s, while ALOHA's are substantially higher (8.09s to 14.89s). Handheld paradigm outperforms ALOHA in data collection efficiency due to its intuitive feedback, while in-situ feedback of \methodname leads to faster data collection.
    \item \textbf{Slip and damage count:} High slip counts indicate poor force and slippage feedback. ALOHA records up to 27 slips (USB plug) and 14 damages (Fragile cup), reflecting its indirect force perception. UMI improves the force feedback and records zero damage events, yet its multi-link transmission still permits noticeable slippage in force-sensitive tasks (e.g., 26 slips during Stamp press). In contrast, \methodname limits slips to at most 2 and records zero damage across all tasks, evidencing its high-fidelity, low-latency tactile feedback.
    \item \textbf{CPUT:} As a combined metric of efficiency and reliability, CPUT highlights the overall advantage of \methodname: it achieves scores from 0.2292 (USB plug) to 0.3356 (Stamp press), substantially outperforming ALOHA and UMI. For tasks like stamp press, the CPUT of \methodname reaches 0.3497, which exceeds ALOHA and UMI by nearly 3 times, confirming its advantage under precise contact requirements.
\end{itemize}
Overall, these results indicate that in-situ design of \methodname improves both the accuracy and efficiency of human-operated data collection in contact-rich tasks.

\textbf{Evaluation of visuo-tactile modality fusion.}
To understand how our policy integrates visual and tactile inputs, we extract and visualize both the decoder's cross-attention and the encoder's self-attention maps during inference. After running the model forward, we compute the cross-attention and self-attention via registered forward hooks.
\begin{figure}[h]
  \centering
  \includegraphics[width=\linewidth]{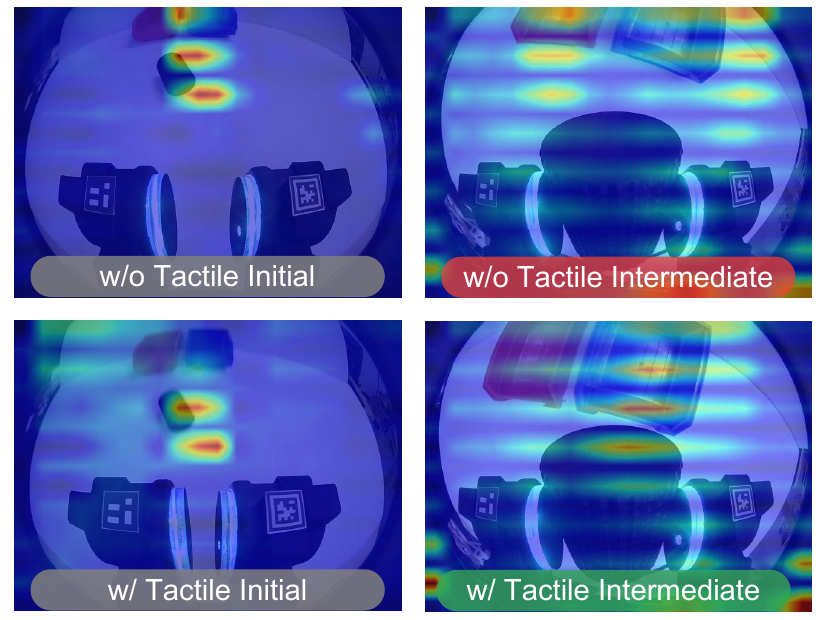} 
    \caption{\textbf{Visual attention heatmaps in texture classification.} Cross‐attention overlays on the visual input:  
    Left two panels (Vision‐only ACT): (a) before grasp and (b) after grasp.  
    Right two panels (Visuo‐tactile ACT): (c) before grasp and (d) after grasp.}
  \label{fig:vis_attn}
\end{figure}
Specifically, the cross-attention from the ACT decoder is captured by averaging across heads, resulting in a single importance vector of length $\mathcal{S}$. This vector is then divided into visual and tactile portions, with each reshaped to its spatial layout and upsampled to the original image resolution for overlay. The encoder self-attention is captured from the attention heads, then the averaged self-attention is reshaped and overlaid onto visual and tactile images.

\begin{figure}[b]
  \centering
  \includegraphics[width=\linewidth]{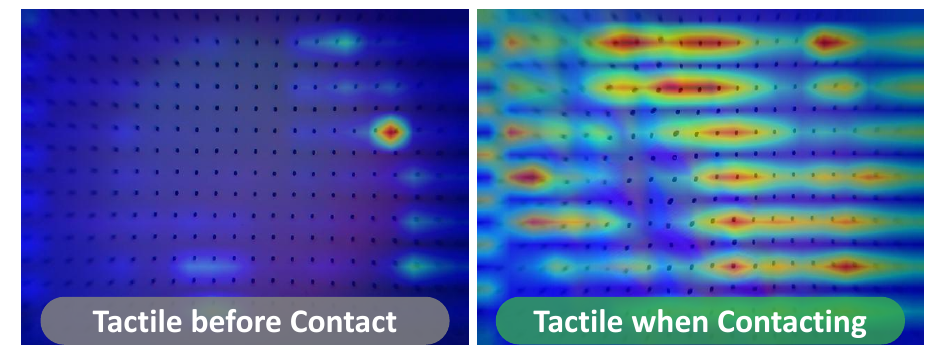} 
  \caption{\textbf{Attention heatmap on tactile images.} We overlay the decoder's cross-attention on tactile images during inference of texture classification. \textbf{Left:} at the initial state, the attention is spread across the gel surface, \textbf{Right:} after contact, the attention focuses on the region of deformation.}
  \label{fig:tac_attn}
\end{figure}
\cref{fig:vis_attn} shows the decoder cross-attention overlays on the visual input during the texture classification task. In the first row, we show two snapshots from the vision-only ACT model: the left image is at the initial state of inference, where attention is correctly focused on the target cylindrical object; the right image is after the gripper has lifted the object, the model's attention is diffusely allocated to both the red and blue bins due to the lack of tactile cues. In the second row, we show the corresponding stages for our model with tactile integration: initial attention in the left image is on the target object again, but after lift (right), the model uses tactile feedback to recognize texture and concentrates its attention exclusively on the correct (blue) bin. This comparison demonstrates how tactile information refines the decision of policy during inference.

\cref{fig:tac_attn} depicts how the ACT model's decoder cross-attention shifts over the tactile image during inference of texture classification. The left image shows the attention map just before the gripper makes contact: since no deformation has happened, attention is broadly distributed over the gel layer. In contrast, the right image captures the moment immediately after contact, where the model concentrates its attention on the region of deformation that corresponds to the object's surface texture. This dynamic reallocation of attention enables the policy to detect fine-grained tactile cues and infer both texture and object geometry in real time.

\end{appendices}

\end{document}